\documentclass[acmsurvey]{acmart}
\AtBeginDocument{%
  }

\setcopyright{acmlicensed}
\copyrightyear{2025}
\acmYear{2025}
\acmDOI{XXXXXXX.XXXXXXX}

\acmJournal{POMACS}
\acmVolume{37}
\acmNumber{4}
\acmArticle{111}
\acmMonth{8}

\usepackage{multirow}
\usepackage{makecell}
\usepackage{longtable}
\usepackage{subfig}
\usepackage{amsmath}
\usepackage{booktabs}
\usepackage{tikz}
\usetikzlibrary{mindmap}
\usepackage{tabularx}
\usepackage{ragged2e}
\usepackage{caption}

\begin{document}

\title{A Survey of Event Causality Identification: Taxonomy, Challenges, Assessment, and Prospects}

\author{Qing Cheng}
\authornote{Both authors contributed equally to this research.}
\email{chengqing@nudt.edu.cn}
\author{Zefan Zeng}
\authornotemark[1]
\email{zengzefan@nudt.edu.cn}
\orcid{0000-0002-4125-0669}
\affiliation{%
  \institution{Laboratory for Big Data and Decision, National University of Defense Technology}
  \city{Changsha}
  \state{Hunan}
  \country{China}
}

\author{Xingchen Hu}
\affiliation{%
  \institution{Laboratory for Big Data and Decision, National University of Defense Technology}
  \city{Changsha}
  \state{Hunan}
  \country{China}}
\email{xhu4@ualberta.ca}

\author{Yuehang Si}
\affiliation{%
   \institution{Laboratory for Big Data and Decision, National University of Defense Technology}
  \city{Changsha}
  \state{Hunan}
  \country{China}}
  \email{siyuehang@nudt.edu.cn}

\author{Zhong Liu}
\affiliation{%
 \institution{Laboratory for Big Data and Decision, National University of Defense Technology}
  \city{Changsha}
  \state{Hunan}
  \country{China}}
  \email{liuzhong@nudt.edu.cn}

\renewcommand{\shortauthors}{Cheng et al.}

\begin{abstract}
Event Causality Identification (ECI) has become an essential task in Natural Language Processing (NLP), focused on automatically detecting causal relationships between events within texts. This comprehensive survey systematically investigates fundamental concepts and models, developing a systematic taxonomy and critically evaluating diverse models.
We begin by defining core concepts, formalizing the ECI problem, and outlining standard evaluation protocols. Our classification framework divides ECI models into two primary tasks: Sentence-level Event Causality Identification (SECI) and Document-level Event Causality Identification (DECI). For SECI, we review models employing feature pattern-based matching, machine learning classifiers, deep semantic encoding, prompt-based fine-tuning, and causal knowledge pre-training, alongside data augmentation strategies. For DECI, we focus on approaches utilizing deep semantic encoding, event graph reasoning, and prompt-based fine-tuning. Special attention is given to recent advancements in multi-lingual and cross-lingual ECI, as well as zero-shot ECI leveraging Large Language Models (LLMs). We analyze the strengths, limitations, and unresolved challenges associated with each approach.
Extensive quantitative evaluations are conducted on four benchmark datasets to rigorously assess the performance of various ECI models. We conclude by discussing future research directions and highlighting opportunities to advance the field further.

\end{abstract}



\received{20 February 2024}
\received[revised]{12 March 2025}
\received[accepted]{5 June 2025}

\maketitle

\section{Introduction}\label{sec:intro}
With the continuous growth of big data, the channels and methods for acquiring unstructured text are constantly expanding. One of the primary challenges now lies in how to automatically extract valuable information and knowledge from these texts. This has become a significant area of research in Natural Language Processing (NLP). Events form the core content of texts, and various research directions have emerged around events, such as Event Extraction (EE) \cite{ee}, Event Relation Extraction (ERE), and Event Coreference Resolution (ECR) \cite{ekgsurvey}. In recent years, Event Causality Identification (ECI) has gained increasing attention as an important and challenging task \cite{liu-erisurvey}. As a critical subtask of ERE, ECI aims to predict whether there is a causal relationship between given events in a text. ECI has been widely applied in tasks such as question answering systems \cite{eci-qa}, information retrieval \cite{eci-ir}, event prediction \cite{eci-ep}, knowledge graph construction \cite{eci-kgc}, and reading comprehension \cite{eci-rc}.

\begin{figure}
    \centering
    \includegraphics[width=0.7\columnwidth]{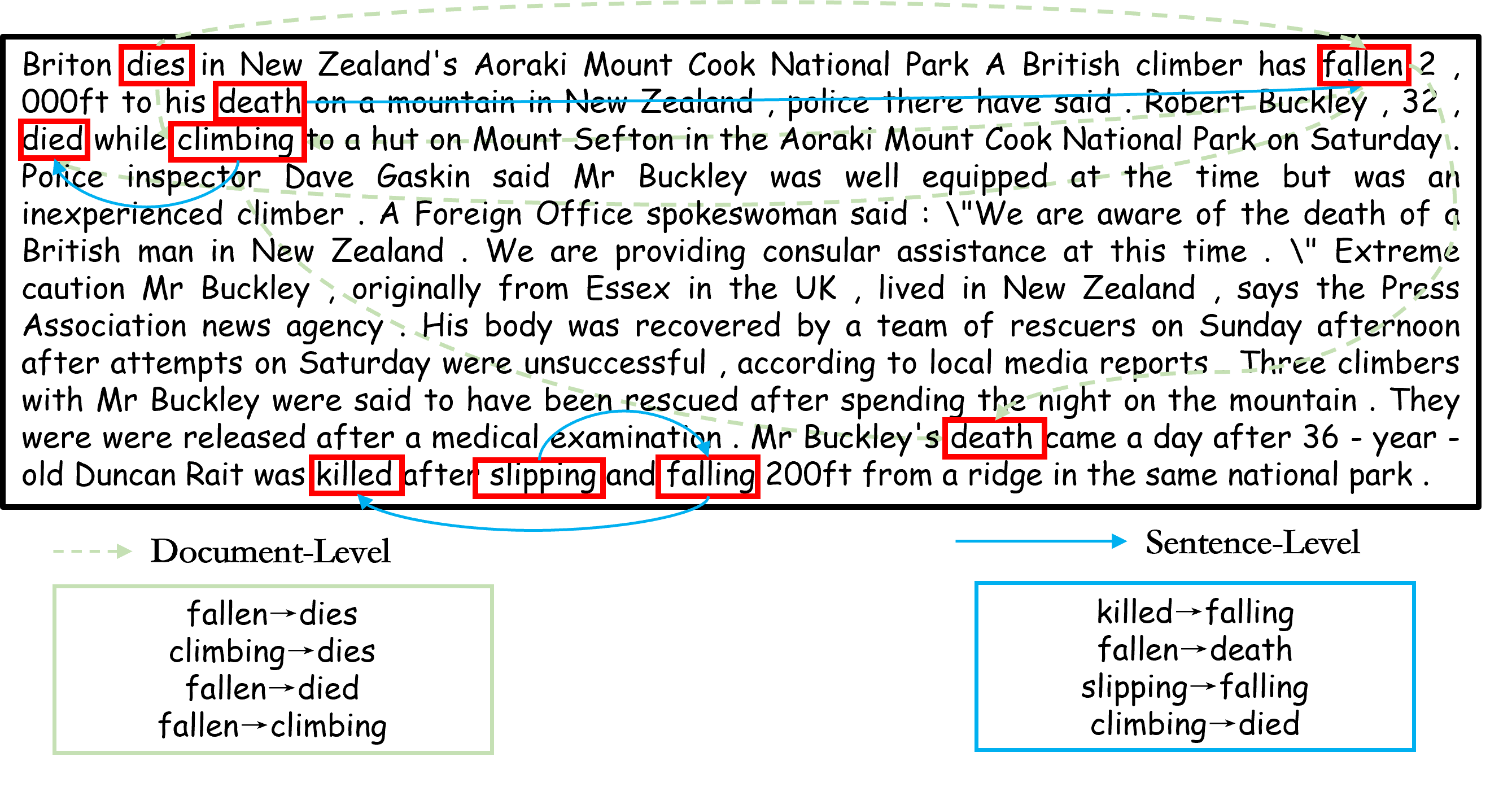}
    \caption{An example of ECI. The red boxes indicate event mentions. The blue solid arrows represent intra-sentence event causalities, while the green dashed arrows represent inter-sentence event causalities.}
    \label{fig:example}
\end{figure}

In the ECI task, events are represented by their triggers, known as "event mentions." The task then becomes determining which event mentions in a given text have a causal relationship. Figure \ref{fig:example} provides an example of event causalities in text. ECI is vital for text understanding and decision-making, as it uncovers the causes and effects of events, helping to analyze risks and opportunities for more informed, data-driven decisions. This capability is particularly essential in domains that require complex reasoning, such as finance \cite{finance}, law \cite{law}, healthcare \cite{medical}, and the military \cite{military}.

ECI focuses solely on extracting specific types of relations—cause and effect. However, compared to the general ERE task, causality identification is more challenging due to several factors:
\begin{enumerate}
    \item \textbf{Implicitness}:  Causal links are often implied rather than directly stated, requiring deeper contextual and semantic understanding.
    \item \textbf{Long-distance dependencies}: Causalities may span multiple sentences or paragraphs, requiring models to capture distant interactions.
    \item \textbf{Complexity}: Causal chains can involve multi-hop links (e.g., "earthquake→collapsed→trapped"), adding complexity to identification.
    \item \textbf{Sample imbalance}: In supervised settings, negative samples (non-causal pairs) dominate, challenging models to learn effectively from fewer positive cases.
    \item \textbf{Limited labeled training data}: With most available datasets being small in scale, models face the dual challenge of accurately parsing semantic relationships while extracting generalizable causal patterns from restricted annotations.
     \item {\textbf{Confused by associations and hidden confounders}: Event causalities are easily confused with correlational associations or influenced by hidden common causes (latent factors affecting both events), leading to causal hallucination. }
\end{enumerate}

ECI research bifurcates into two distinct paradigms based on event pair localization: (1) \textit{Sentence-level ECI (SECI)}, which analyzes explicit causal relations within single sentences—a mature field with 7-8 years of methodological development; and (2) \textit{Document-level ECI (DECI)}, an emerging domain (post-2022) addressing cross-sentence causal networks requiring modeling of implicit signals and discourse-level dependencies.

Early ECI models relied on feature pattern recognition, using cues like lexical signals \cite{feature1,feature2,feature3}, temporal features \cite{mirza-2014,mirza-tonelli-2014-analysis}, and co-occurrences \cite{esc,do-etal-2011-minimally}. With the rise of Deep Learning (DL), more advanced methods emerged, enabling models to better capture contextual information by deeply encoding text semantics \cite{eresurvey-acm}. Transformer-based \cite{transformer} Pre-trained Language Models (PLMs) \cite{bert} have revolutionized ECI, as they are trained on large corpora, enhancing semantic understanding and producing high-quality event and context representations, thereby improving causal identification \cite{bert-multiano,semsin,knowmmr}. Since 2023, Large Language Models (LLMs) have become widely popular. Leveraging massive datasets and self-supervised learning at scale, modern LLMs have achieved unprecedented knowledge representation capabilities, demonstrating remarkable few-shot and zero-shot task adaptability through enhanced contextual understanding. \cite{chatgpt-eci,llm-eci, dreci}. This has advanced research in event extraction, relation extraction, and question answering. However, research by Gao et al. \cite{chatgpt-eci} highlights that while LLMs can perform zero-shot ECI with simple prompts, they are prone to "causal hallucination," leading to many false positives. Hence, even though LLMs have significantly advanced tasks related to textual event analysis, ECI remains highly challenging.


Several surveys have examined event causality in text and ERE. Asghar \cite{asghar2016automaticextractioncausalrelations} reviewed early rule-based and statistical methods for causal extraction but did not cover deep learning (DL) approaches. Zhao et al. \cite{zhao-survey} and Xie et al. \cite{xie-eresurvey} provided overviews of advances in ERE, with the former summarizing general trends in entity causality and the latter comparing DL-based supervision methods. Liu et al. \cite{liu-erisurvey} focused on causal and temporal relationships but with limited coverage of recent methods. Yang et al. \cite{yang-survey} and Ali et al. \cite{ali-survey} reviewed explicit and implicit causality extraction, limited to work prior to 2021. Liu et al. \cite{liu-survey2} covered methods, challenges, and datasets for event relation identification but lacked detailed method classification and in-depth analysis for ECI models.

In this paper,we present a systematic review of ECI research. First, we establish a comprehensive conceptual taxonomy framework based on core concepts, fundamental techniques, and methodological advances in ECI. Second, we conduct rigorous performance evaluations through quantitative comparisons of existing approaches. Finally, we identify emerging trends and future research directions in this rapidly evolving field. We also highlighting representative applications across key domains. Our main contributions are detailed as follows:
\begin{itemize}
    \item We detail several concepts of ECI, including problem formalization, datasets, evaluation protocols, and key technologies.
    \item {We present the first comprehensive taxonomy for ECI, as shown in Table \ref{tab:taxonomy}. This framework systematically classifies existing models based on their technical characteristics and modeling approaches. It also serves as a theoretical foundation for ECI methodologies, providing guidance for the design of future models.}
    \item {We perform a quantitative comparison of different ECI methods on four benchmark datasets.}
    \item We discuss future directions in ECI, highlighting key challenges and potential solutions for advancing the field.
    \item {We introduce typical applications of ECI in critical domains like healthcare, finance, and social science.}
\end{itemize}

\begin{table}[t]
  \centering
  \caption{{Taxonomy of ECI models, distinguishing SECI and DECI subtasks. All models/methods are labeled as ``\textit{Model Name}/Author Name + Reference Index'' for clarity.}}
  \begin{scriptsize}
  \renewcommand{\arraystretch}{1.2}
  \setlength{\tabcolsep}{3pt}
  \begin{tabularx}{\linewidth}{|>{\Centering}m{1.8cm}|>{\Centering}m{2cm}|>{\Centering}m{2cm}|>{\Centering}m{1.8cm}|>{\Centering}X|}
    \hline
    \textbf{Research Task} & \textbf{Classification} & \textbf{Subcategory} & \textbf{Sub-subcategory} & \textbf{Reference Literature} \\
    \hline
    \multirow{16}[36]{*}{\makecell{Sentence-Level\\Event\\Causality\\Identification}} 
      & \multirow{3}[4]{*}{\makecell{\\Feature Pattern-\\Based Matching}} 
      & \makecell{\\Template Matching} & \multirow{3}[6]{*}{--} 
      & Khoo et al. \cite{khoo_many_2002}, Chan et al. \cite{feature1}, Khoo et al. \cite{feature2}, Rehbein et al. \cite{template_catching}, Khoo et al. \cite{khoo-etal-2000-extracting} \\
    \cline{3-3}\cline{5-5}
      & & \makecell{\\Syntactic Patterns} & 
      & Ishii and Ma \cite{ishiiandma,ishii_incremental_2012}, Caselli and Vossen \cite{esc}, Girju and Moldvan \cite{feature3}, Girju \cite{girju-2003-automatic}, Ittoo and Bouma \cite{Ittoo-1}, Ittoo et al. \cite{Ittoo-2}, Radinsky et al. \cite{radinsky} \\
    \cline{3-3}\cline{5-5}
      & & Co-occurrence Patterns & 
      & Do et al. \cite{do-etal-2011-minimally}, Caselli and Vossen \cite{esc}, Luo et al. \cite{luo_et_al} \\
    \cline{2-5}
      & \multirow{2}[6]{*}{\makecell{\\Machine Learning-\\Based Classification}} 
      & \makecell{\\Explicit Lexical\\-Based Methods} & \multirow{2}[4]{*}{\makecell{\\ \\ --}}
      & Chang and Choi \cite{chang_and_choi}, Mirza and Tonelli \cite{mirza-tonelli-2014-analysis,mirza-tonelli-2016-catena}, Mirza \cite{mirza-2014}, Hidey and Mckeown \cite{hidey}, Pakray and Gelbukh \cite{pakray}, Bethard et al. \cite{Bethard2008}, Hashimi et al. \cite{hashimy}, Riaz and Girju \cite{riaz-girju-2014-recognizing} \\
    \cline{3-3}\cline{5-5}
      & & \makecell{\\Implicit Feature\\-Based Methods} & 
      & Zhao et al. \cite{zhao_event_2016}, Girju et al. \cite{GIRJU2010589}, Blanko et al. \cite{blanco-etal-2008-causal}, Bollegala et al. \cite{BOLLEGALA2018}, Qiu et al. \cite{QIU20171623}, Bethard and Martin \cite{bethard_and_marting}, Rink et al. \cite{Rink2010}, Ning et al. \cite{ning-etal-2018-joint}, Yang and Mao \cite{yang_and_mao} \\
    \cline{2-5}
      & \multirow{6}[6]{*}{\makecell{\\Deep\\Semantic\\Encoding}} 
      & Simple Encoding & {--}
      & Choubey and Huang \cite{choubey-huang-2017-sequential} \\
    \cline{3-5}
      & & \multirow{3}[1]{*}{\makecell{\\Textual\\Information\\Enhancement}} 
      & Semantic Structure 
      & \textit{Tri-CNN} \cite{tri-cnn}, Ayyanar et al. \cite{ayyanar}, \textit{CEPN} \cite{cepn}, \textit{SemSIn} \cite{semsin}, \textit{SemDI} \cite{semdi} \\
    \cline{4-5}
      & & & \multirow{2}[1]{*}{Event Information}
      & Ponti and Korhonen \cite{ponti-korhonen-2017-event}, \textit{BERT+MA} \cite{bert-multiano}, \textit{DualCor} \cite{dualcor}, \textit{Siamese BiLSTM} \cite{siamese} \\
    \cline{4-5}
      & & & Label Information 
      & \textit{SCL} \cite{scl}, \textit{RPA-GCN} \cite{rpa-gcn}, \textit{ECLEP} \cite{eclep}, \textit{CF-ECI} \cite{cfeci} \\
    \cline{3-5}
      & & \multirow{2}[1]{*}{\makecell{\\External\\Knowledge\\Enhancement}} 
      & \multirow{2}[0]{*}{\makecell{Representation \\ Enhancement}} 
      & \textit{KnowMMR} \cite{knowmmr}, \textit{DSET} \cite{dset}, \textit{ECIFF} \cite{eciff}, \textit{ECIHS} \cite{ecihs}, \textit{KLPWE} \cite{klpwe}, \textit{DFP} \cite{dfp}, \textit{KIGP} \cite{kigp} \\
    \cline{4-5}
      & & & \multirow{2}[0]{*}{\makecell{Reasoning \\ Enhancement}} 
      &  \textit{K-CNN} \cite{k-cnn}, \textit{BERT+MFN} \cite{bert+mfn}, \textit{LSIN} \cite{lsin}, Chen and Mao \cite{chenandmao}, \textit{C3Net} \cite{c3net}, \textit{GCKAN} \cite{gckan} \\
    \cline{2-5}
      & \multirow{2}[2]{*}{\makecell{Prompt-Based\\Fine-Tuning}} 
      & \multirow{2}[0]{*}{\makecell{Masked Prediction \\ Methods}} & \multirow{2}{*}{\makecell{\\ --}}
      & \textit{MRPO} \cite{mrpo}, \textit{KEPT} \cite{kept}, \textit{DPJL} \cite{dpjl}, \textit{ICCL} \cite{iccl}, \textit{HFEPA} \cite{hfepa}, \textit{RE-CECI} \cite{receci} \\
    \cline{3-3}\cline{5-5}
      & & Generative Methods & 
      & \textit{GenECI} \cite{geneci}, \textit{GenSORL} \cite{gensorl} \\
    \cline{2-5}
      & \multicolumn{3}{c|}{Causal Knowledge Pre-Training} 
      & \textit{CausalBERT} \cite{causalbert}, \textit{Causal-BERT} \cite{causal-bert}, \textit{UnifiedQA} \cite{UnifiedQA} \\
    \cline{2-5}
      & \multirow{2}[2]{*}{\makecell{Data\\Augmentation}} 
      & External Knowledge-Based Augmentation & \multirow{2}{*}{\makecell{\\ --}}
      & \textit{KnowDis} \cite{knowdis}, \textit{LearnDA} \cite{learnda}, \textit{CauSeRL} \cite{causerl} \\
    \cline{3-3}\cline{5-5}
      & & Generation-Based Augmentation & 
      & \textit{ERDAP} \cite{erdap}, \textit{PCC} \cite{pcc}, \textit{EDA} \cite{eda}, \textit{BT-ESupCL} \cite{BT-ESupCL} \\
    \hline
    \multirow{8}{*}{\makecell{Document-Level\\Event\\Causality\\Identification}} 
      & \multicolumn{3}{c|}{Deep Semantic Encoding} 
      & \textit{KADE} \cite{kade}, \textit{SENDIR} \cite{sendir}, \textit{DiffusECI} \cite{diffuseci}, Ensemble Learning \cite{ensemble} \\
    \cline{2-5}
      & \multicolumn{3}{c|}{\makecell{\\Event Graph Reasoning-Based Methods}} 
      & \textit{DCGIM} \cite{dcgim}, \textit{RichGCN} \cite{richgcn}, \textit{GESI} \cite{gesi}, \textit{ERGO} \cite{ergo}, \textit{CHEER} \cite{cheer}, \textit{DocECI} \cite{doceci}, \textit{EHNEM} \cite{ehnem}, \textit{PPAT} \cite{ppat}, \textit{iLIF} \cite{ilif}, Gao et al. \cite{ilp} \\
    \cline{2-5}
      & \multicolumn{3}{c|}{Prompt-Based Fine-Tuning} 
      & \textit{CPATT} \cite{cpatt}, \textit{DAPrompt} \cite{daprompt}, \textit{HOTECI} \cite{hoteci} \\
    \cline{2-5}
    & \multicolumn{3}{c|}{\makecell{\\Large Language Model-Based Methods}}
      & Gao et al. \cite{llm-eci}, \textit{KLop} \cite{KLop}, \textit{LKCER} \cite{lkcer}, \textit{KnowQA} \cite{knowqa}, Zhang et al. \cite{zhang_llm}, Wang et al. \cite{wang-etal-2024-event-causality}, \textit{Dr. ECI} \cite{dreci} \\
    \cline{2-5}
    & \multicolumn{3}{c|}{\makecell{\\Multi-lingual and Cross-Lingual ECI} }
      & Lai et al. \cite{meci}, Zhu et al. \cite{zhu2023chinese}, \textit{GIMC} \cite{gimc}, \textit{PTEKC} \cite{ptekc}, \textit{Meta-MK} \cite{metamk} \\
    \hline
  \end{tabularx}
        
  \end{scriptsize}
  \label{tab:taxonomy}
\end{table}

The remainder of this paper is organized as follows: Section \ref{sec:background} presents the relevant concepts of ECI, reviews commonly used datasets, and evaluation metrics. Section \ref{sec: keytechs} introduces key technologies for ECI. Sections \ref{sec: seci} and \ref{sec: deci} provide a comprehensive overview of the classification framework for SECI and DECI, summarizing the core techniques of various models, and evaluating their strengths and limitations. {Section \ref{generalizable} examines advancements in generalizable ECI, encompassing cross-lingual ECI and zero-shot ECI, which have emerged as research hotspots over the past two years. Section \ref{sec: assess} presents performance assessment and quantitative evaluation (Appendix \ref{appendix}) of the performance of existing classical methods on four benchmark datasets.} Section \ref{sec:challenge} outlines future research directions for ECI. {We also present typical applications and practical implications of ECI across three key domains: healthcare, finance, and social science sectors in Appendix \ref{application}.}
 
\section{Background}\label{sec:background}
In this section, we begin by defining key concepts, including events, event causality, and ECI. We then provide a comprehensive review of the benchmark datasets commonly used for extracting event causalities in recent years. Lastly, we outline the evaluation metrics employed to assess the performance of ECI models.
\subsection{Definitions and Problem Formalization}
\subsubsection{Event}
An “event” is an objective fact defined by specific individuals, objects, and actions at a certain time and place, typically stored in unstructured formats like text, images, and videos. This study focuses on events contained within textual data. 

For example, the sentence "\textit{Around 8:40 AM on February 6, shortly after a subway train left a station in Moscow, the second carriage suddenly exploded, and the train caught fire.}" describes two events: "\textit{the train exploded after departure}" and "\textit{the train caught fire.}" Textual events often include elements like triggers and arguments. To facilitate analysis, triggers are used as \textbf{event mentions}. In this example, "\textit{exploded}" and "\textit{fire}" serve as the trigger words for the respective events.
\subsubsection{Event Causality}
Broadly speaking, causalities between events refer to the driving connections between them. If two events, A and B, have a causality represented as A→B, it means that the occurrence of A lead to the occurrence of B, with A being the cause and B being the effect. 
In this context, the causality between events in a text can be defined as follows:

\begin{definition}
\textbf{Event Causality}. For a given context \( \mathcal{D} \), causality \( e_i \rightarrow e_j \) between two events \( e_i, e_j \in \mathcal{D} \) exists if and only if they satisfy at least one of the following three conditions:
\begin{enumerate}
     \item \textbf{Enablement/Prevention}: $e_i$ occurs before $e_j$ and has effect on (\textit{enables or prevents}) the occurrence of $e_j$;
    \item \textbf{Counterfactual Dependence}: \( e_j \) \textit{would not occur} if \( e_i \) did not occur (\textit{counterfactual necessity});
    \item \textbf{Deterministic Implication}: The occurrence of \( e_i \) \textit{necessarily entails} the occurrence of \( e_j \) (\textit{logical sufficiency}).
\end{enumerate}
\end{definition}
\textbf{Example 1:}

Text: "A car bomb explosion$_{e_1}$ in Algeria resulted in 11 deaths$_{e_2}$ and 31 injuries$_{e_3}$."  

Causalities: explosion→death, explosion→injury

\textbf{Example 2:  }

Text: "The earthquake$_{e_1}$ killed$_{e_2}$ 14 people, hundreds of people trapped$_{e_3}$ in collapsed$_{e_4}$ buildings."  

Causalities:  earthquake→killed, earthquake→collapsed, collapsed→trapped, earthquake→trapped 

\textbf{Example 3:  }

Text: "1. Warship INS Sukanya on Thursday foiled$_{e_1}$ a piracy attempt in the Gulf of Aden between Somalia and Yemen. 2. The ship was escorting$_{e_2}$ a group of five merchant vessels through the IRTC when the bid$_{e_3}$ occurred."  

Causalities: foiled→escorting, foiled→bid

In Example 1, the word "resulted" clearly indicates the causality between the events "explosion" and "death" as well as "injury," which is referred to as "\textbf{explicit} causality." In contrast, Example 2 lacks explicit connecting words or cues to indicate causalities, leading to three pairs of "\textbf{implicit} causality." 

In the first two examples, all events are contained within the same sentence, termed "\textbf{intra-sentence} causality" or "\textbf{sentence-level} causality." In Example 3, the events "foiled," "escorting," and "bid" are not in the same sentence, creating what is known as "\textbf{inter-sentence} causality" or "\textbf{document-level} causality."

\subsubsection{Event Causality Identification}
Event causality identification is a subtask within event relation extraction, aimed specifically at identifying the causalities between designated events. ECI focuses on extracting specific causal relationships, such as \textit{cause} or \textit{caused\_by}. Based on the definitions of ERE in the literature \cite{ekgsurvey}, event causality identification can be formally defined as follows:
\begin{definition}
\textbf{Event Causality Identification}. Given a text \( \mathcal D = [w_1, \ldots, w_n] \) containing \( n \) words and event sets \( \mathcal E = \{e_1, e_2, \ldots, e_k\} \) along with their mentions \( m_1, m_2, \ldots, m_k \subset \mathcal D \). For any two events \( e_i,e_j \in \mathcal E \), determine their relationship type as one of \{\textbf{cause, caused\_by, None}\}.
\end{definition}
{
Beyond classifying ECI into SECI and DECI, depending on whether the text spans multiple languages, ECI tasks can also be classified into \textit{monolingual} ECI and \textit{multi-lingual} ECI. Additionally, based on the availability of labeled samples for learning, ECI tasks can be divided into \textit{supervised} ECI, \textit{semi-supervised} ECI, and \textit{unsupervised} ECI (\textit{zero-shot} ECI).}

\subsection{Datasets}\label{dataset}
Annotated datasets are critical to advancing research in ECI. At present, several datasets are either specifically crafted for ECI tasks or can be adapted to evaluate ECI performance. Here, we provide a curated overview of 10 widely adopted datasets for ECI evaluation, detailing each dataset's source, version, and statistics.
\begin{itemize}
    \item \textbf{SemEval-2010 Task 8} \cite{semeval2010}: An enhanced version of SemEval-2007 \cite{semeval2007} for classifying semantic relations between noun pairs, featuring nine relations (including causality) and an "OTHER" category. It emphasizes contextual clarity and reduces category overlap.
     \item \textbf{Chinese Event Causality (CEC}) \cite{cec}: A Chinese causality dataset, extracted using layered Conditional Random Fields (CRFs). It includes many-to-many, explicit, inter-sentence, cross-sentence, cross-paragraph, embedded, and overlapping causalities.
    \item \textbf{Causal-TimeBank (CTB)} \cite{ctb}: Based on the TempEval-3 corpus \cite{tempeval3}, CTB tags events in TimeML (Temporal Markup Language) format. It uses $\mathtt {CLINK}$ tags for causal links between events and $\mathtt{C-SIGNAL}$ tags for words or phrases that indicate causality, with a focus on explicit causalities.
    \item \textbf{CaTeRS} \cite{CATERs}: Based on 320 five-sentence stories from the ROCStories corpus \cite{rocstories}, this dataset covers everyday scenarios with 13 types of causal and temporal relationships, focusing on narrative structure and story-driven causality.
    \item \textbf{BECauSE Corpus 2.0} \cite{becausecorpus}: An updated BECauSE corpus with texts from \textit{The Wall Street Journa}l, \textit{The New York Times}, and the \textit{Penn Treebank}. It includes detailed annotations of causal language and seven other semantic relations often linked with causality, focusing on explicit causalities.
    \item \textbf{EventStoryLine Corpus (ESL)} \cite{esc}: Drawn from 22 topics in the ECB+ dataset on events like disasters, shootings, and trials, ESC supports cross-document reasoning and narrative generation. Available versions include v0.9, v1.0, v1.5, and v2.1. The most widely used version is v0.9.
    \item \textbf{FinReason} \cite{finreason}: A Chinese corpus in the financial domain aimed at extracting reasons for significant events in public company announcements. Each document may include multiple structured events, each with zero, one, or more reasons.
    \item \textbf{MAVEN-ERE} \cite{maven}: The first large-scale, human-annotated ERE dataset, using documents from various topics on $\texttt {Wikipedia}$\footnote{\url{https://simple.wikipedia.org/}}. It builds on \cite{ogorman-etal-2016-richer} with an enhanced annotation scheme to support various relationship types, including event coreference, temporal, causal, and sub-event relationships.
    \item \textbf{MECI} \cite{meci}: The first multi-lingual dataset for event causality identification, sourced from $\texttt {Wikipedia}$ in five languages (English, Danish, Spanish, Turkish, and Urdu). It offers valuable resources for multi-lingual recognition and cross-lingual transfer learning.
    \item {\textbf{COPES} \cite{copes}: The first benchmark dataset for contextualized commonsense causal reasoning, constructed from temporal event sequences in ROCStories \cite{rocstories}. It provides carefully annotated cause-effect pairs within narrative contexts, enabling research on implicit causal relationships in everyday scenarios.}
\end{itemize}
{Among the aforementioned datasets, ESL and CTB are the most widely adopted, while MECI and MAVEN-ERE have gained increasing popularity in recent years.}
Table \ref{tab:datasets} summarizes the statistics of these datasets. In addition to the commonly used datasets mentioned above, there are also smaller, annotated at the sentence granularity, less frequently used, or non-open-source event causality datasets, such as Richer Event Descriptions \cite{richer}, The Causal News Corpus (CNC) \cite{cnc}, Penn Discourse Treebank 2.0 \cite{penn}, and CausalityEventExtraction \footnote{\url{https://github.com/liuhuanyong/CausalityEventExtraction}}, which are not detailed here. {To facilitate experimentation and comparison, we have organized these datasets and made them available on $\mathtt{Gitee}$\footnote{\url{https://gitee.com/KGAWS/eci-related-works-and-datasets}}.
}

\begin{table*}[htbp]
\centering
\caption{Statistics of Common ECI Datasets, "*" indicates Chinese datasets, "$\Delta$" indicates multi-lingual datasets. COPES adopts event-sequence-based annotation, thus are listed separately in the table.}
\label{tab:datasets}
\begin{small}
\begin{tabular}{|c|c|c|c|c|c|}
\hline
\textbf{Datase} & \textbf{Document Count} & \textbf{Event Count} & \textbf{Relation Count} & \textbf{Causality Count} & \textbf{Release Year} \\
\hline
SemEval-2010 Task 8 & 10,717 & 9,994 & 10,717 & 1,331 & 2010 \\
\hline
CEC* & 332 & 5,954 & 340 & 340 & 2011 \\
\hline
Causal-TimeBank & 184 & 6,813 & 7,608 & 318 & 2014 \\
\hline
CaTeRS & 320 & 2,708 & 2,715 & 700 & 2016 \\
\hline
BECauSE Corpus 2.0 & 121 & 2,386 & 1,803 & 1,803 & 2017 \\
\hline
Event StoryLine (v0.9) & 258 & 5,334 & 5,655 & 5,519 & 2017 \\
\hline
FinReason* & 8,794 & 12,861 & 11,006 & 11,006 & 2021 \\
\hline
MAVEN-ERE & 4,480 & 112,276 & 1,290,050 & 57,992 & 2022 \\
\hline
MECI$^{\Delta}$ & 3,591 & 46,000 & 11,000 & 11,000 & 2022 \\
\hline
\end{tabular}
\end{small}
\end{table*}

\subsection{Evaluation Metrics}
As with most classification tasks, existing research commonly evaluates the performance of ECI using three metrics: precision, recall, and F1-score. Precision is defined as the percentage of true positives (causal event pairs) among all samples classified as positive by the model. Recall is calculated as the ratio of correctly identified positives to the total number of true positives. The F1-score, which is the harmonic mean of precision and recall, provides a balanced measure of precision and recall, evaluating the overall performance of causality identification.

Equations \ref{eq:eval1}–\ref{eq:eval3} present the formulas for these three evaluation metrics:

\begin{equation}\label{eq:eval1}
    \text{precision} = \frac{|\mathcal P^t|}{|\hat{\mathcal P|}},
\end{equation}
\begin{equation}
    \text{recall} = \frac{|\mathcal P^t|}{|\mathcal P|},    
\end{equation}
\begin{equation}\label{eq:eval3}
    \text{F1-score} = \frac{2 \times \text{precision} \times \text{recall}}{\text{precision} + \text{recall}},    
\end{equation}
where \( \mathcal P^t \) represents the set of correctly identified causal event pairs, \( \hat{\mathcal P} \) is the set of all identified causal event pairs, and \( \mathcal P \) is the set of all annotated causal event pairs.

\section{Key Technologies}\label{sec: keytechs}
This section covers four essential techniques for ECI: syntactic parsing, pattern matching, text embedding, and graph embedding. These techniques encompass all current methods for ECI, serving as the foundation for implementing and understanding related models. 

\subsection{Syntactic Parsing}
Syntactic parsing analyzes sentence grammar to build syntactic trees or dependency graphs, revealing structural relationships between subjects, predicates, and objects. It includes constituent parsing, using Context-Free Grammar $G = (V, \Sigma, R, S)$ to form phrase-based parse trees, and dependency parsing, which creates trees showing direct word relationships, with semantic dependency parsing (e.g., Abstract Meaning Representation) capturing deeper meanings via graphs. While effective for explicit causal links with minimal training, it struggles with implicit causality, generates redundant links, and depends heavily on parser accuracy, especially for complex sentences.

\subsection{Pattern Matching}
Pattern matching uses predefined rules to identify segments that fit specific patterns, such as fixed strings, regular expressions, or syntactic structures. Patterns are defined based on the application needs, then matched with input text or syntax-parsed data. Regular expressions and part-of-speech (POS) tags are handled by annotation engines, while syntactic matching relies on graph-based similarity.
\subsection{Text Embedding}

Text embedding converts language elements into continuous vector spaces, capturing semantics for ECI. It includes word-level, contextual, and graph embeddings. Word-level embeddings map words to vectors, with static types (e.g., Word2Vec \cite{word2vec}, GloVe \cite{glove}) using fixed vectors and dynamic types (e.g., BERT, RoBERTa \cite{roberta}) generating context-aware vectors via self-attention, adaptable through fine-tuning. For a sentence $X = [x_1, ..., x_n]$, encoding yields:
\begin{equation}\label{eq:eventemb}
    \mathbf{X} = [\mathbf{x}_0, \mathbf{x}_1, ..., \mathbf{x}_n] = \text{Encode}([x_1, ..., x_n]),
\end{equation}
where $\mathbf{x}_0$ is the sentence embedding, and event mention $e = [x_m, ..., x_{m+l}]$ has embedding:
\begin{equation}
    \mathbf{e} = \frac{1}{l+1} \sum_{i=0}^{l} \mathbf{x}_{m+i}.
\end{equation}
Contextual embeddings, using CNNs \cite{cnn}, RNNs \cite{lstm}, or bidirectional RNNs \cite{bilstm}, capture sentence-level features, while Transformer-based PLMs \cite{bert,roberta} model long-range dependencies via self-attention. Graph embeddings represent causality graphs $\mathcal{G} = (\mathcal{V}, \mathcal{E})$, mapping nodes (events) and edges (causalities) into vectors using GNNs like GCN \cite{gcn}, GAT \cite{gat}, or GraphSAGE \cite{graphsage}. GNNs update node representations via:
\begin{equation}
\mathbf{h}_v^{(l+1)} = \text{UPDATE}^{(l)}(\mathbf{h}_v^{(l)}, \text{AGGREGATE}^{(l)}({\mathbf{h}_u^{(l)} | u \in \mathcal{N}(v)})).
\end{equation}
Graph embedding models enhance ECI by capturing complex relationships and filtering noise in long texts.

\section{Sentence-Level Event Causality Identification}\label{sec: seci}
Sentence-level event causality is typically expressed through specific lexical items and syntactic structures, characterized by numerous explicit cues and prototypical causal patterns. Owing to the limited contextual requirements of SECI, its semantic interpretation is relatively straightforward. 

{We classify current SECI methods into five categories: feature pattern-based matching, machine learning-based classification, deep semantic encoding, prompt-based fine-tuning, and causal knowledge pre-training. Additionally, we discuss research on data augmentation strategies that aim to address the challenge of insufficient training samples in supervised ECI.}
\subsection{Feature Pattern-Based Matching}
\begin{figure}[t]
    \centering
    \includegraphics[width=0.7\columnwidth]{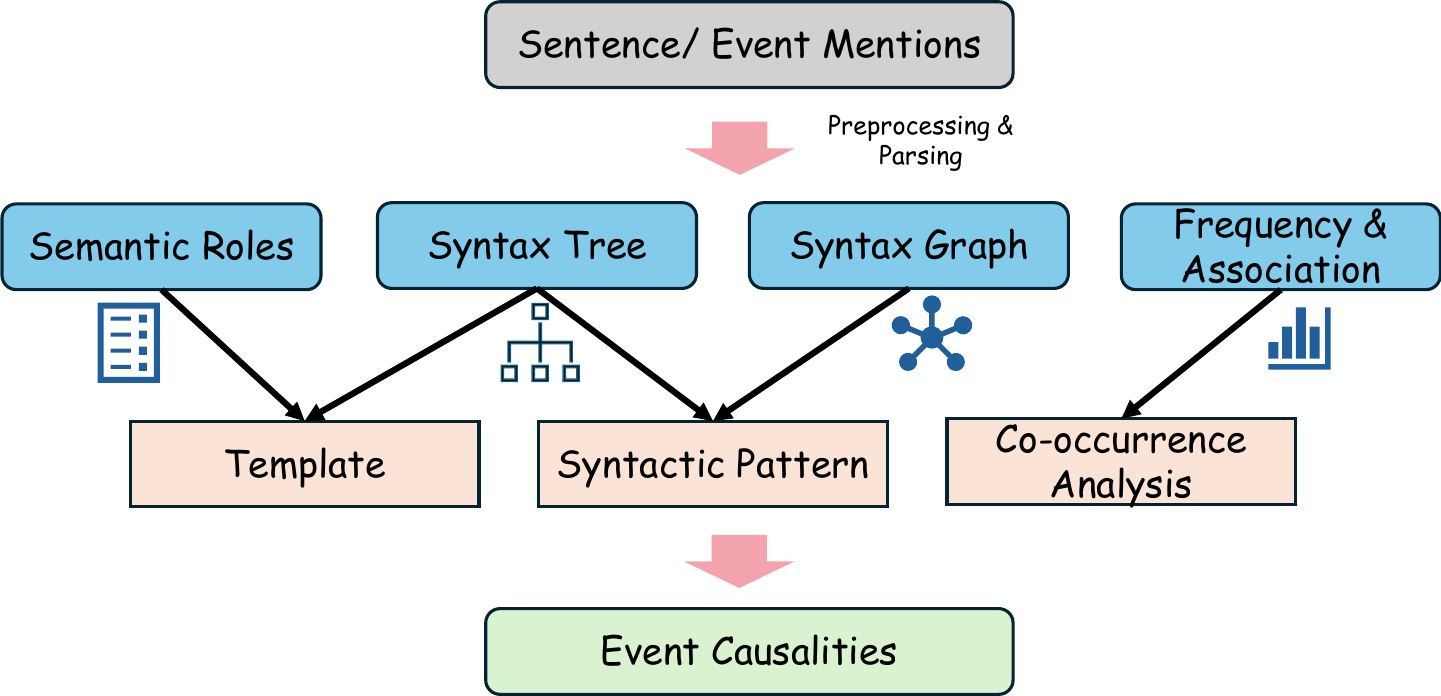}
    \caption{{The feature pattern-based matching framework preprocesses and parses input sentences and events to derive semantic or syntactic features or computes frequency and statistical associations, which serve as the basis for pattern matching.}} 
    \label{fig:featurepattern}
\end{figure}

Before deep learning became prevalent, feature pattern-based matching was the main approach for SECI. These methods leverage lexical cues, syntax, and pattern recognition to detect causal patterns. {This approach includes three types: template matching, syntactic patterns, and co-occurrence. The general framework of feature pattern-based matching methods is shown in Figure \ref{fig:featurepattern}.}

\subsubsection{Template matching-based methods}
\begin{figure}[t]
    \centering
    \includegraphics[width=0.95\columnwidth]{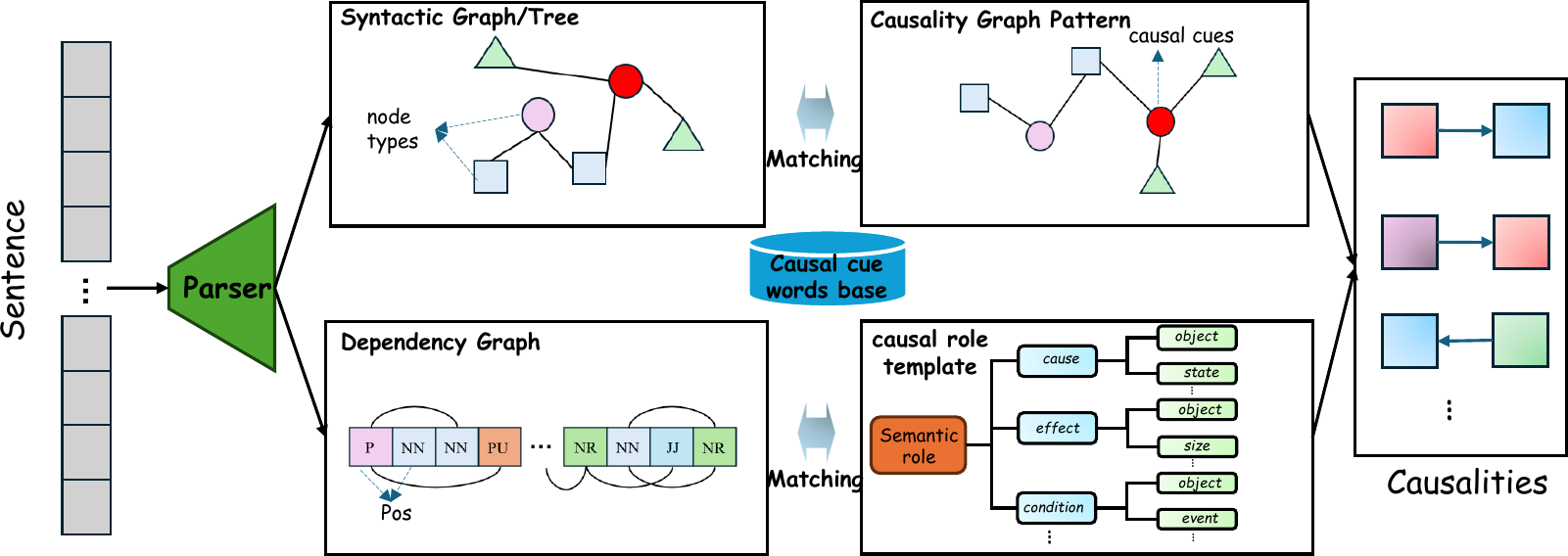}
    \caption{The general framework of template matching methods. Dependency graph or syntax graph/tree of the sentence is generated using a parser, which is then matched against predefined templates to identify causalities.} 
    \label{fig:template}
\end{figure}

Template matching-based methods identify causalities using predefined patterns, such as role or graph templates (see Figure \ref{fig:template}). Khoo et al. \cite{khoo_many_2002} provided an in-depth analysis of causality templates and structures. Khoo et al. \cite{feature2} also outlined five templates for causal patterns: (1) causal conjunctions, (2) causative verbs, (3) resultative complements, (4) conditional sentences, and (5) adverbs/adjectives with causal meaning, achieving a 68\% identification rate on test data. Chan et al. \cite{feature1} proposed constructing templates that represent roles and attributes within causal contexts. When text matches a template, matched words populate template slots to represent causalities. 
{Rehbein et al. \cite{template_catching} proposed a German causal relation extraction method using a lexicon and annotated corpus of 1,000+ verb-based templates.}
Khoo et al. \cite{khoo-etal-2000-extracting} enhanced accuracy and scalability by using syntactic graph templates, which align with parse trees to pinpoint cause and effect segments in text, making template matching more precise. Template matching methods are interpretable and sensitive to explicit causalities. However, they can only identify causalities that fit predefined templates, making it challenging to address implicit, complex, or irregular expressions of causality.

\subsubsection{Syntactic pattern-based methods} Syntactic pattern-based methods identify causalities using predefined structures. Ishii and Ma \cite{ishiiandma,ishii_incremental_2012} developed a topic-event causal model, representing text as a causal network graph, merging similar events based on keywords and scoring importance. Caselli and Vossen \cite{esc} proposed the Order Presentation (OP) method, pairing events by their sequence to infer causality (also for DECI). Girju and Moldvan \cite{feature3} semi-automatically identified causal syntactic patterns by extracting noun phrase pairs from $\texttt{WordNet}$ \cite{wordnet} and matching them to sentences with the pattern "$\texttt{NP1 verb NP2}$." {Girju \cite{girju-2003-automatic} automated causality detection using syntactic patterns from parsed text, refined through dependency analysis and graph optimization. Ittoo and Bouma \cite{Ittoo-1} modeled events as a causality graph, using shortest paths for syntactic patterns and scoring reliability. Ittoo et al. \cite{Ittoo-2} employs $\texttt{Wikipedia}$ as a knowledge base to first extract reliable causal patterns through minimally supervised learning, then transfers these patterns to match event pairs in target domain corpora. Radinsky et al. \cite{radinsky} extracted causality patterns from historical news to create causal rules, using ontologies to predict future events.}
These methods detect complex causalities, automatically generate patterns with parsing tools, and resolve ambiguities. However, reliance on parsers increases computational demands and may reduce performance if parsers are inaccurate.

\subsubsection{Co-occurrence pattern-based methods} Co-occurrence pattern-based methods identify causalities based on event co-occurrence frequency and semantic associations. Do et al. \cite{do-etal-2011-minimally} proposed a minimally supervised learning approach that uses pointwise mutual information and inverse document frequency to measure relationships between events. They calculate co-occurrence counts to predict causalities and leverage discourse connectives for additional context. Integer Linear Programming (ILP) is then used to ensure causal consistency. {Luo et al. \cite{luo_et_al} relied on Pointwise Mutual Information (PMI) to identify causal relationships by harvesting a term-based causality co-occurrence network from a large web corpus.} Caselli and Vossen \cite{esc} employed Positive Pointwise Mutual Information (PPMI) to evaluate the semantic relevance of event pairs by analyzing co-occurrence frequencies in large datasets. The improved PPMI-CONTAINS model adds temporal constraints, requiring event pairs to have PPMI values within a specified range and to share the same time frame. Co-occurrence-based methods use statistical information from large corpora instead of predefined templates, enabling them to address implicit causalities and adapt to different domains. However, reliance on statistical associations can result in many false positives.

\subsubsection{Open Challenges} Pattern matching-based methods are simple, efficient, and interpretable, with low computational demands, making them popular in early research. However, they lack the ability to capture deep semantic information and rely heavily on external parsers and predefined lexical or syntactic patterns. This dependence limits their scalability, flexibility, and generalizability, causing performance to degrade significantly with complex semantics, unfamiliar structures, or noisy text.

\begin{figure}[t]
    \centering
    \includegraphics[width=0.75\columnwidth]{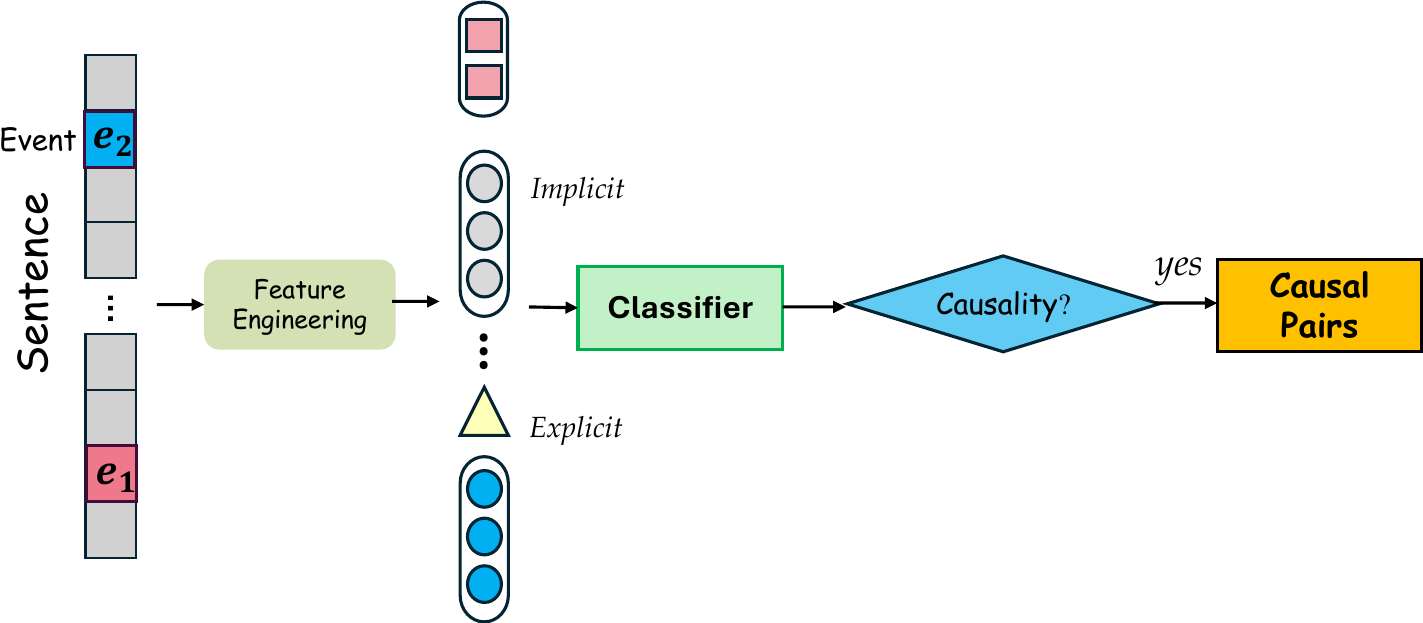}
    \caption{{General framework of machine learning-based classification for ECI. These methods typically begin by extracting explicit or implicit causal features from text, then employ machine learning models to train classifiers for determining the existence of causal relations and identifying the involved events/entities.} }
    \label{fig:mlbased}
\end{figure}

{
\subsection{Machine Learning-based Classification}}
{
This approach employs machine learning to model either explicit cue words or implicit features associated with causal relations, then builds a classifier to determine causality between event pairs. This category mainly includes explicit lexical-based methods and implicit feature-based methods. The general framework of this approach is shown in Figure \ref{fig:mlbased}.}

\subsubsection{{Explicit Lexical-based Methods}}
{
The approach detects explicit causal vocabulary features and learns diagnostic lexical patterns to identify causalities. Chang and Choi \cite{chang_and_choi} used lexical pair and cue phrase probabilities with a Naive Bayes classifier to improve causality detection. Mirza and Tonelli \cite{mirza-tonelli-2014-analysis} inferred causal chains via causal verbs and dependencies, employing an SVM-based model \cite{mirza-tonelli-2016-catena} with lexical-syntactic features and dependency paths. Mirza \cite{mirza-2014} integrated temporal rules and lexical-syntactic features for classification. Hidey and Mckeown \cite{hidey} used parallel $\texttt{Wikipedia}$ corpora to identify causal markers, training a classifier with parallel corpus and lexical semantic features. Pakray and Gelbukh \cite{pakray} applied a decision tree with $\texttt{WordNet}$, dependency parsing, and positional features to detect cause-effect relations. Bethard et al. \cite{Bethard2008} used SVM classifiers with lexical and syntactic features from event pairs and context. Hashimi et al. \cite{hashimy} classified ontological causalities by analyzing semantic relations and context. Riaz and Girju \cite{riaz-girju-2014-recognizing} combined shallow linguistic features with semantics of $\texttt{FrameNet}$ and $\texttt{WordNet}$, using ILP to identify causal relations. These models learn diverse explicit feature combinations but struggle with sparse annotated data and capturing deep semantic features.
}

\subsubsection{{Implicit Feature-based Methods}}
{
Implicit feature-based methods identify latent causal indicators using contextual patterns, distributional semantics, and deep linguistic features to detect indirect causal signals. Zhao et al. \cite{zhao_event_2016} measured sentence similarity with causal connectives, using a Naive Bayes model for causality detection. Girju et al. \cite{GIRJU2010589} developed a supervised semantic relation classification system with multi-level features and an SVM classifier. Blanko et al. \cite{blanco-etal-2008-causal} used machine learning to classify causality-encoding syntactic patterns. Bollegala et al. \cite{BOLLEGALA2018} detected adverse drug reactions from social media with lexical patterns and a supervised classifier. Qiu et al. \cite{QIU20171623} applied CRFs with temporal features for sequence labeling in typhoon scenarios. Bethard and Martin \cite{bethard_and_marting} used machine learning with $\texttt{WordNet}$ and Google N-gram features for causality identification. Rink et al. \cite{Rink2010} employed a graph-based approach with lexical-syntactic-semantic subgraph patterns for binary classification. Ning et al. \cite{ning-etal-2018-joint} proposed a joint inference model with a perceptron algorithm and ILP for temporal and causal relation extraction. Yang and Mao \cite{yang_and_mao} integrated a multi-level relation extractor with extended lexical-semantic features for causality classification. These methods enhance causality detection by capturing indirect semantics but require large, high-quality datasets and can be computationally complex for real-time or resource-limited applications. This approach also remains limited in extracting profound causalities.
}
\subsubsection{{Open Challenges}}
{
Compared with simplistic template-based methods, machine learning-based classification methods offer greater flexibility by combining diverse lexical-syntactic-semantic features and statistical learning, achieving better coverage of implicit causality. However, they often require manual feature engineering and may lack generalization capability for complex causal patterns, as they primarily rely on predefined linguistic features rather than learning hierarchical representations automatically.
}
\subsection{Deep Semantic Encoding}
With advancements in deep learning, neural sequence models have enabled richer representations for text and events, particularly in managing long-distance dependencies. PLMs generate context-sensitive, dynamic word embeddings through large-scale pre-training, thereby capturing more intricate semantics. 

Deep semantic encoding-based ECI methods employ deep sequence models or PLMs as encoders, applying tailored encoding strategies to capture the semantic and contextual information of events and generate high-dimensional embeddings. The embeddings of two target events are concatenated and input into a classifier (e.g., a multilayer perceptron), which learns the features of these high-dimensional embeddings through neural networks to assess the presence of causality. { Existing deep semantic encoding-based ECI models can be categorized into three types according to their encoding strategies: simple encoding, textual information enhanced encoding, and external knowledge enhanced encoding.}

\subsubsection{Simple Encoding} Simple encoding strategies leverage sequence models (e.g., RNNs, LSTMs) and PLMs (e.g., BERT, RoBERTa, and Longformer \cite{longfromer}) as encoders to obtain contextual embeddings of event pairs, which are then input into a classifier. Although few of them are specifically covered in dedicated literature \cite{choubey-huang-2017-sequential}, these methods are frequently used as baselines for comparison in research.

For sequence models, an input sentence \( X = [x_1, \dots, x_n] \) is processed by first obtaining static word embeddings for each word, followed by encoding with a sequence model to capture contextual information, generating event embeddings as follows:
\begin{equation}
    \mathbf x_i=\texttt{Emb}(x_i),
\end{equation}
\begin{equation}
    \widetilde{\mathbf{X}}=\left[{\widetilde{\mathbf{x}}}_1,{\widetilde{\mathbf{x}}}_2,\ldots,{\widetilde{\mathbf{x}}}_n\right]=\texttt{SeqModel}\left(\left[\mathbf{x}_1,\mathbf{x}_2,\ldots,\mathbf{x}_n\right]\right),
\end{equation}
\( \texttt{Emb} \) and \( \texttt{SeqModel} \) denote static encoding and sequence model encoding, respectively. The event embeddings are calculated based on the words corresponding to event mentions. For PLMs, event embeddings are derived using equation \ref{eq:eventemb}.

The embeddings of two events, \( \mathbf{e}_i \) and \( \mathbf{e}_j \), are combined to form the event pair representation using various composition strategies. Common approaches include concatenation (e.g., \( \mathbf{ep}_{ij} = [\mathbf{e}_i; \mathbf{e}_j] \) or \( \mathbf{ep}_{ij} = [\mathbf e_{\text{[CLS]}}; \mathbf{e}_i; \mathbf{e}_j] \)), element-wise subtraction (e.g., \( \mathbf{ep}_{ij} = \mathbf{e}_i - \mathbf{e}_j \)), and element-wise multiplication (e.g., \( \mathbf{ep}_{ij} = \mathbf{e}_i \odot \mathbf{e}_j \)). This concatenated vector \( \mathbf{ep}_{ij} \) is then input to a classifier, typically a Multi-Layer Perceptron (MLP), as follows:
\begin{equation}\label{simpreason}
    \mathbf{p}=\mathrm{softmax}(\mathrm{MLP}(\mathbf{ep}_{ij}))=\mathrm{softmax}(f(\mathbf {W\cdot ep}_{ij}+\mathbf b)),
\end{equation}
\( \mathbf{W} \) and \( \mathbf{b} \) are learnable parameter matrices, , $f$ is the activation function, and \( \mathbf{p} \in \mathbb{R}^2 \) (ignoring direction) or \( \mathbf{p} \in \mathbb{R}^3 \) (considering direction). Each dimension of \( \mathbf{p} \) corresponds to the probability of the causality or non-causality.

During training, parameters are optimized on an annotated dataset using cross-entropy loss to enable causality classification:
\begin{equation}
    \mathcal{L} = -\sum_{j=1}^{N} \sum_{i=1}^{C} y_i^j \log\left(p_i^j\right),
\end{equation}
where \( C = 2 \) or \( 3 \) corresponds to the output dimension of \( \mathbf{p} \), \( y_i^j \) represents the true label, \( p_i^j \) the predicted probability, and \( N \) the number of labeled event pairs. { Since the annotation of ECI datasets are often imbalanced, adaptive focal loss is frequently employed as an alternative, which addresses class imbalance by down-weighting well-classified examples and focusing training on hard negatives:
\begin{equation}
    \mathcal{L}_{focal} = -\sum_{j=1}^{N} \sum_{i=1}^{C} \alpha_i^j (1-p_i^j)^\gamma y_i^j \log\left(p_i^j\right),
\end{equation}
where \( \gamma \) modulates the focusing intensity and \( \alpha_i^j \in [0,1] \) is a class-balancing factor. Encoder parameters can also be updated as needed.}

Simple encoding methods demonstrate limited capacity in capturing event contextual information and expressing causal semantic features. They require large amounts of labeled data for training, are prone to false positives, and exhibit relatively weak cross-domain generalization.

\begin{figure}[t]
    \centering
    \includegraphics[width=0.9\columnwidth]{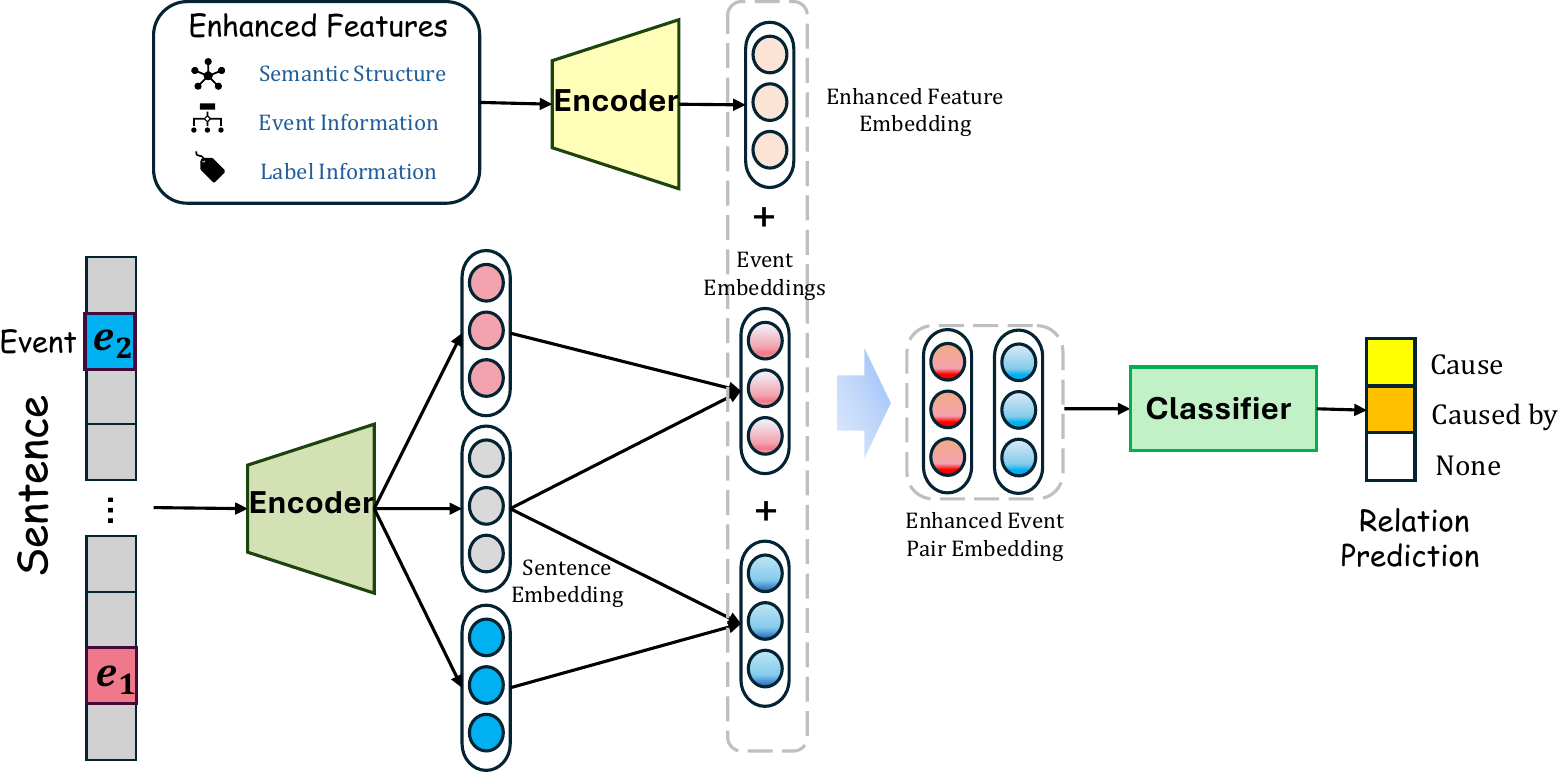}
    \caption{Framework of textual information enhanced encoding for ECI. Text and event mentions, along with additional features (such as semantic structure, event information, and labels), are encoded as embeddings and fused. A classifier is then used to generate the predicted probability for the event relation type.} 
    \label{fig: enhanced_feat}
\end{figure}

\subsubsection{{Textual Information Enhanced Encoding}}
{
Textual information enhanced encoding strategies leverage sequence models or PLMs combined with rich feature in the text to capture contextual information, improving the quality of event and event-pair representations.} These strategies integrate additional information into embeddings before classification (see Figure \ref{fig: enhanced_feat}). Current research include three main enhancement approaches: semantic structure enhancement, event information enhancement, and causal label enhancement. Table \ref{tab:enhanced_features} summarizes the enhancement details of these encoding models.

\textbf{Semantic structure enhancement} encoding strategies use structured semantic information, such as dependency relations and semantic graphs, to enrich event representations. \textit{Tri-CNN} \cite{tri-cnn} integrates syntactic and contextual information via a three-channel CNN, combining paths and event-pair embeddings for global representations. {Ayyanar et al. \cite{ayyanar} used grammatical tags and parse tree distances to enhance CNN-based event encoding. \textit{CEPN} \cite{cepn} leverages positional indices and sequence generation to address multiple and overlapping causalities.} \textit{SemSIn} \cite{semsin} encodes events with BERT, parses text into AMR-based semantic graphs, and aggregates information using GNNs. \textit{SemDI} \cite{semdi} uses a unified encoder to capture semantic dependencies, leveraging PLMs and a Cloze analyzer to query causalities via a causality discriminator. These strategies can handle complex sentences but increase training and inference times.

\textbf{Event information enhancement} encoding strategies enrich event representations by capturing semantics, location, time, and participants. {Ponti and Korhonen \cite{ponti-korhonen-2017-event} proposed features for semantics and positional information to enhance FFNN-based event encoding. Kadowaki et al. \cite{bert-multiano} used BERT to encode event roles and positions, training multiple classifiers to combine annotator perspectives for predictions.} \textit{DualCor} \cite{dualcor} employs a dual grid tagging scheme with BERT to capture event correlations and distinguish cause-effect pairs. Yang et al. \cite{siamese} use a Bi-LSTM-based Siamese network to incorporate event triggers and arguments (e.g., time, location), inferring causalities via a knowledge graph-inspired approach. These methods enrich event representations but are computationally intensive and require well-designed event features.

\textbf{Label information enhancement} encoding strategies refine event pair embeddings using causal labels through contrastive learning, graph interactions, or counterfactual reasoning. {\textit{RPA-GCN} \cite{rpa-gcn} uses entity annotation and entity-location-aware GAT to enhance relational feature perception.} \textit{ECLEP} \cite{eclep} builds an event-pair interaction graph, integrating causal labels with GAT-updated representations. \textit{SCL} \cite{scl} constructs triplets with causal labels, using contrastive loss for distinctive embeddings. \textit{CF-ECI} \cite{cfeci} applies counterfactual reasoning to reduce bias, adjusting predictions with causal labels via weighted debiasing. These methods maintain accuracy in noisy or ambiguous contexts but rely heavily on extensive labeled data.

\begin{table}[htbp]
  \centering
  \caption{{Enhanced Features of Different Textual Information Enhanced Encoding-based Models}}
  \begin{small}
  \begin{tabular}{|c|c|c|c|}
    \hline
    \textbf{Enhance Strategy} & \textbf{Model} & \textbf{Encoder} & \textbf{Enhance Features} \\
    \hline
    \multirow{5}{*}{Semantic structure} 
    & Tri-CNN \cite{tri-cnn} & RCNN & Syntactic information, contextual information \\
    \cline{2-4}
    & SemSIn \cite{semsin} & BERT/GNN/BiLSTM & Event-centric structure, event association structure \\
    \cline{2-4}
    & Ayyanar et al. \cite{ayyanar} & CNN & Grammatical tags, parse tree distances \\
    \cline{2-4}
    & CEPN \cite{cepn} & Positional LSTM & Structural indices, sequence generation \\
    \cline{2-4}
    & SemDI \cite{semdi} & RoBERTa & Semantic dependency \\
    \hline
    \multirow{4}{*}{Event information} 
    & Ponti and Korhonen \cite{ponti-korhonen-2017-event} & FFNN & Event semantics, positional features \\
    \cline{2-4}
    & BERT+MA \cite{bert-multiano} & BERT & Event location, text segment, features (roles) \\
    \cline{2-4}
    & DualCor \cite{dualcor} & BERT & Event type and arguments \\
    \cline{2-4}
    & Siamese BiLSTM \cite{siamese}& BiLSTM & Event time, location, participants \\
    \hline
    \multirow{4}{*}{Label information} 
    & SCL \cite{scl} & BERT & Labeled positive and negative samples \\
    \cline{2-4}
    & RPA-GCN \cite{rpa-gcn} & BERT/GAT & Entity-location annotations \\
\cline{2-4}
    & ECLEP \cite{eclep} & BERT/GAT & Causally related event annotations \\
    \cline{2-4}
    & CF-ECI \cite{cfeci} & BERT & Causal labels \\
    \hline
  \end{tabular}
  \end{small}
  \label{tab:enhanced_features}
\end{table}

\subsubsection{{External Knowledge Enhanced Encoding}}
\begin{figure}[t]
    \centering
    \includegraphics[width=0.95\columnwidth]{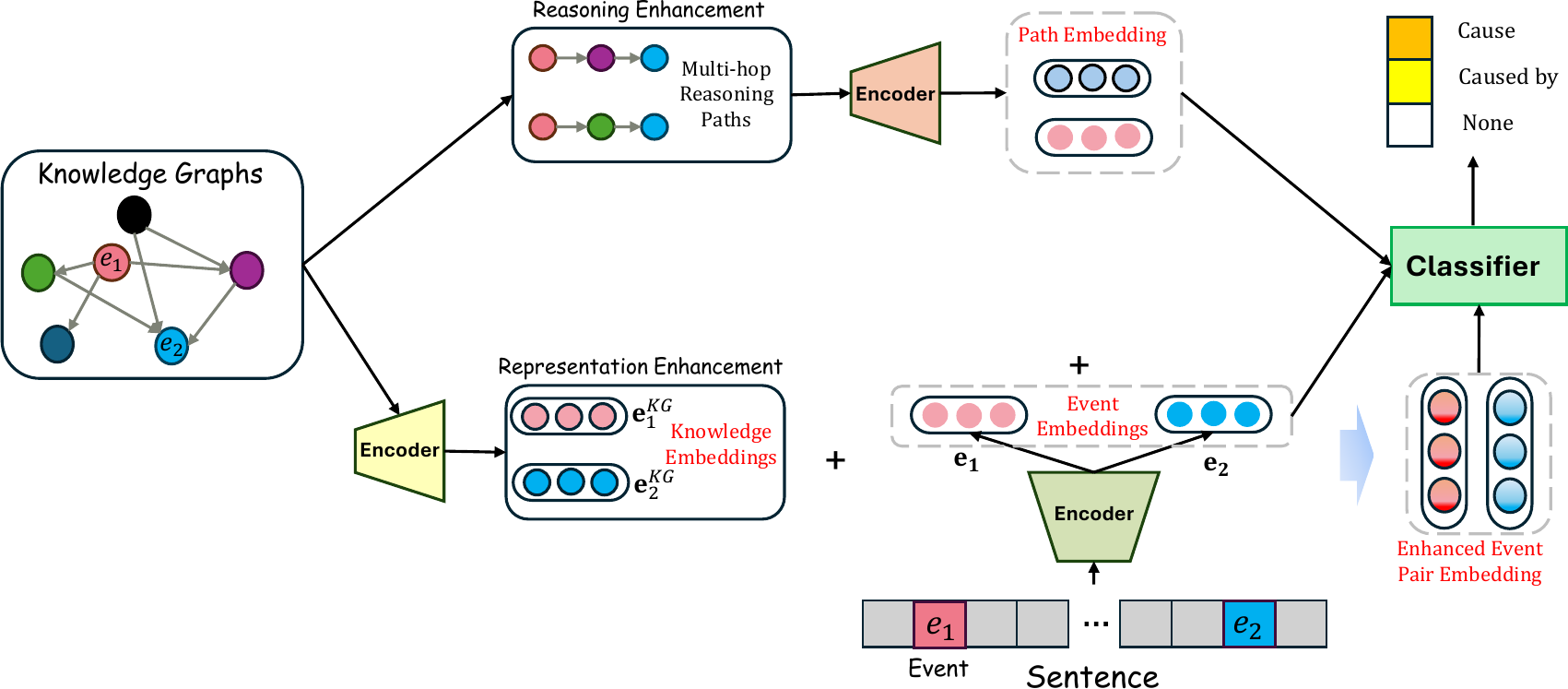}
    \caption{Framework of external knowledge enhanced encoding for ECI. KGs generate supplementary annotated data enhance model training. Events in KGs are encoded via graph embeddings and merged with contextual embeddings of event mentions, improving representation quality. Additionally, KG subgraphs and reasoning paths support causal reasoning.} 
    \label{fig: external}
\end{figure}

In ECI, causal expressions also vary across topics and contexts, limiting model performance and increasing overfitting risk. To address this, recent studies have integrated external knowledge, primarily through Knowledge Graphs (KGs), which provide causal background knowledge to alleviate data scarcity and support reasoning (see Figure \ref{fig: external}). {External knowledge enhanced encoding methods can be categorized into representation enhancement and reasoning enhancement.}

\textbf{Representation enhancement} methods improve event pair representations by integrating external knowledge and contextual information using PLMs or GNNs. \textit{KnowMMR} \cite{knowmmr} enhances ECI generalization by retrieving $\texttt{ConceptNet}$ knowledge, encoding it with BERT, and applying event masking for context-independent patterns. \textit{DSET} \cite{dset} uses temporal prompts and $\texttt{ConceptNet}$ to enhance event representations, with RoBERTa encoding knowledge and GraphSAGE capturing event graph dependencies. \textit{ECIFF} \cite{eciff} combines contextual, semantic, and syntactic features, encoding $\texttt{CauseNet}$ \cite{causenet} causal triples and using adversarial learning for robust event pair embeddings. \textit{ECIHS} \cite{ecihs} captures syntactic, semantic, and background knowledge from $\texttt{ConceptNet}$, using Transformers, syntax trees, AMR graphs, and Graph Transformer \cite{graphtransformer} for enhanced representations. \textit{KLPWE} \cite{klpwe} integrates KG embeddings, morphological, POS, and BERT-based dynamic embeddings to predict causal roles. \textit{DFP} \cite{dfp} fuses commonsense and event semantics via GAT for KG information and BERT for text, pre-training on Masked Language Modeling (MLM), concept triple completion, and text-graph contrastive learning. {\textit{KIGP} \cite{kigp} builds an event knowledge interaction graph, using GCNs to extract interaction features and enhance implicit causality identification.} These methods improve generalization but rely on external knowledge quality. Poorly integrated knowledge may introduce noise, reducing inference accuracy.

\textbf{Reasoning enhancement} methods use external knowledge to construct relational graphs or causal chains, aiding implicit causality reasoning through multi-hop or logical mechanisms. \textit{K-CNN} \cite{k-cnn} integrates $\texttt{WordNet}$ and $\texttt{FrameNet}$ to capture linguistic clues via semantic features and causal scores. \textit{BERT+MFN} \cite{bert+mfn} combines BERT for word-level semantics and MFN for relational reasoning with external knowledge. \textit{LSIN} \cite{lsin} uses $\texttt{ConceptNet}$ and $\texttt{COMET}$ \cite{comet} to build descriptive and multi-hop relational graphs, applying variational inference with BERT-encoded events. Chen and Mao \cite{chenandmao} integrate $\texttt{FrameNet}$ \cite{framenet} and $\texttt{COMET}$ for complex causality, using GCN-based event pair graphs and causality-related words. \textit{C3Net} \cite{c3net} retrieves causal paths from external knowledge, processing them through association, review, and inference for reliable reasoning. \textit{GCKAN} \cite{gckan} builds event graphs with $\texttt{ConceptNet}$, using GAT and contrastive learning to enhance robustness. These methods improve cross-domain causal understanding. However, over-reliance on external knowledge may limit handling of ambiguous causalities, and unreliable reasoning paths can cause error propagation.

\subsubsection{{Open Challenges}}
{Simple encoding strategies are easy to implement and train but are only suited for clear, low-noise texts within a single domain. Textual information enhanced encoding strategies provide a more comprehensive modeling of event causality, especially in complex contexts and long-distance dependencies. They are effective in handling complex semantic structures and exhibit stronger cross-domain generalizability. External knowledge enriches context and reduces overfitting. However, deep semantic encoding-based methods require a substantial amount of labeled data. Without sufficient supervised learning, these models struggle to learn high-quality encoding and classification capabilities, which impacts inference accuracy. Moreover, when using external knowledge for enhancement, model performance relies on the knowledge base's coverage and quality. Insufficient or conflicting knowledge can mislead causal inference. Integrating external knowledge may also introduce noise or redundancy, increasing complexity.}

\subsection{{Prompt-Based Fine-Tuning}}
{
Prompt-based fine-tuning methods utilize the semantic understanding and extensive knowledge of PLMs to identify causalities by employing fixed or dynamic prompt templates. By framing ECI as a masked prediction, multiple-choice, or text generation task, these methods enable PLMs to directly generate causal labels. These methods can be technically categorized into two paradigms: masked prediction approaches and generative methods. Table \ref{tab: prompt} summarizes the prompt templates of these methods.}

\begin{table*}[t]
  \centering
  \caption{Prompt templates and PLMs of prompt-based fine-tuning for ECI.}
    \begin{small}
    \begin{tabular}{|c|c|c|}
    \hline
    Model & Prompt Template & Notations  \\
    \hline
    MRPO \cite{mrpo} &    \makecell{From context [\textit{D}], it is [MASK1] that \\ \textit{E1} has a causal relationship to \textit{E2}.} &  \multirow{14}{*}{\makecell{ [Es][Et][/Es][/Et]: boundary markers. \\ $\mathcal T_a (d_i )$: the i-th demonstration example template. \\ $\mathcal T_p (q)$: prompt template for the query event pair. \\ $[MASK1]$ $\in$ \{Yes,No\}\\ $[MASK2]$ $\in$ \{Cause,Causedby,None\} \\ $[MASK3]$: causal cue words \\  $[MASK4]$: event mention}} \\
    \cline{1-2}
    KEPT \cite{kept} &    [Es] \textit{E1} [/Es] [MASK2] [Et] \textit{E2} [/Et].  &  \\
    \cline{1-2}
    DPJL \cite{dpjl} &   \makecell{ (1) In this sentence, \textit{E1} [MASK2] \textit{E2}. \\(2) The cue word of \textit{E1} [MASK2] \textit{E2} is [MASK3].\\ (3) According to the sentence,\textit{E1} causes [MASK4],\\ \textit{E1} is caused by [MASK4].} &  \\
    \cline{1-2}
    ICCL \cite{iccl}  &   [CLS] $\mathcal T_a (d_1 )$ [SEP] ... $\mathcal T_a (d_n )$ [SEP] $\mathcal T_p (q)$ [SEP]  &  \\
    \cline{1-2}
    HFEPA \cite{hfepa} & \makecell{There is a [MASK2] relation \\ between [Es] \textit{E1} [/Es] and [Et] \textit{E2} [/Et]?}  & \\
    \cline{1-2}
    RE-CECI \cite{receci}  &  In this sentence, \textit{E1} [MASK2] \textit{E2}.   &  \\
    \cline{1-2}
    GenECI \cite{geneci}  &  Is there a causal relation between \textit{E1} and \textit{E2}?   &   \\
    \cline{1-2}
    GenSORL \cite{gensorl}  &  \makecell{Is there a causal relation \\ between  [Es] \textit{E1} [/Es] and [Et] \textit{E2} [/Et]?}    &  \\
    \hline
    \end{tabular}%
    \end{small}
  \label{tab: prompt}%
\end{table*}%

\subsubsection{Masked Prediction Methods}
These approaches formulate ECI as a fill-in-the-blank task where PLMs predict masked tokens in predefined templates to determine causal relationships. They are particularly effective for capturing localized causal patterns through constrained vocabulary prediction. \textit{MRPO} \cite{mrpo} converts ECI into an MLM task. A predefined template with a $\texttt{[MASK]}$ token is used for causal classification. BERT is employed to perform masked token prediction, an MLP classifier processes the prediction for causality determination. By considering all possible masked token values, this approach reduces dependency on specific lexical cues. 
\textit{KEPT} \cite{kept} augments ECI by integrating external knowledge from $\texttt{ConceptNet}$ \cite{conceptnet}. The framework employs a specialized prompt template that demarcates events using boundary tokens, encodes this structured input via BERT, and selectively attends to relevant relational information from $\texttt{ConceptNet}$ to enhance event representations.
\textit{DPJL} \cite{dpjl} frames ECI as a mask prediction task with templates. DPJL adds two auxiliary prompts for cue word detection and causal event detection, respectively. Joint training on these two tasks is then leveraged for model learning. \textit{ICCL} \cite{iccl} combines prompt-based learning with in-context contrastive learning. The model uses RoBERTa to predict $\texttt{[MASK]}$ for the query template. The contrastive learning framework optimizes causal discrimination by dynamically sampling and contrasting positive (causal) and negative (non-causal) event pairs during training. {\textit{HFEPA} \cite{hfepa} captures features from event mentions and segment-level contexts to enrich event semantics and leverages prompt-aware attetion to model latent event relationships.} \textit{RE-CECI} \cite{receci} implements a retrieval-enhanced framework for Chinese ECI by retrieving examples from a stored database using k-nearest neighbors. These retrieved examples are incorporated into a prompt template. 

\subsubsection{ Generative Methods}
These methods employ sequence-to-sequence architectures to generate causal labels and contextual explanations. They excel at handling complex causal scenarios requiring multi-step reasoning and open-ended expression.
\textit{GenECI} \cite{geneci} employs a generative methodology leveraging the T5 \cite{t5} to predict causal labels and produce contextual words that link events within a predefined template, thereby constructing causal paths as output. Through reinforcement learning, the generated output is refined by optimizing for accuracy, semantic similarity, and structural alignment with the input template. \textit{GenSORL} \cite{gensorl} is an improvement on GenECI, which extracts simplified dependency paths from syntax trees to reduce noise in causality prompts. It applies a policy gradient with baseline for more stable training and includes an auxiliary reward to improve similarity between generated outputs and input sentences. 

\subsubsection{Open Challenges}
Prompt-based fine-tuning effectively harnesses the adaptability of PLMs in low-data scenarios, with flexible prompt designs enhancing cross-domain generalization. Nevertheless, an over-reliance on linguistic patterns may result in false negatives for implicit causality. The quality of prompts is paramount, as complex contexts or noise can compromise model robustness if prompts are poorly constructed. Employing soft prompts or dynamic prompts represents a promising strategy to address these limitations and enhance the efficacy of such methods.

\subsection{Causal Knowledge Pre-training}
Causal knowledge pre-training involves retraining or continuously pre-training PLMs with causal commonsense or domain-specific knowledge, enabling zero-shot, end-to-end ECI using simple prompts and MLM. Causal Knowledge PLMs (CKPLMs) improve causal reasoning by leveraging extensive causal knowledge bases and cross-task training, allowing effective application to ECI tasks with straightforward prompt designs.

\textit{CausalBERT} \cite{causalbert} collects multi-level causal pairs from unstructured text and injects this knowledge into BERT to improve causal reasoning. Its trained in three stages: (1) unsupervised pre-training with a BERT-based masked language model; (2) causal pair classification and ranking with data from sources like  $\texttt{CausalBank}$\footnote{\url{https://github.com/eecrazy/CausalBank/}} and $\texttt{ConceptNet}$; and (3) application to causal classification tasks using simple prompts or fine-tuning for causal reasoning tasks.

\textit{Causal-BERT} \cite{causal-bert} enhances implicit causality detection by using sentence and event contexts. It consists of three frameworks: (1) C-BERT, which encodes events and sentence contexts to detect causality; (2) event-aware C-BERT, which integrates event and sentence contexts for improved predictions; and (3) masked event C-BERT, which replaces events with “BLANK” to learn implicit structures, later fine-tuned with event-aware C-BERT.

\textit{UnifiedQA } \cite{UnifiedQA} is a multi-task Question Answering (QA) model trained on various question formats. The latest version, UnifiedQA-v2, expanded its training data to include 20 datasets with causal knowledge, further improving its generalizability.

\textit{Summary}. CKPLMs enable rapid ECI with minimal labeled data, effectively handling implicit causality and generalizing well across domains. However, challenges like causal hallucination and prompt sensitivity lead to high false positive rates and potential misinformation. When the knowledge in the text conflicts with the knowledge in CKPLMs, it may result in incorrect responses. Furthermore, the model's constrained size results in suboptimal reasoning performance.

\begin{figure}[t]
    \centering
    \includegraphics[width=0.95\columnwidth]{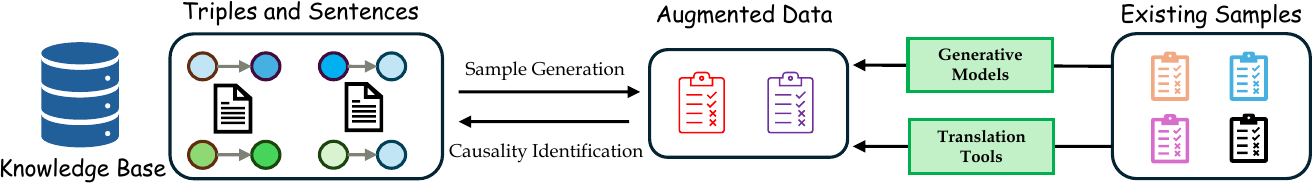}
    \caption{Framework of data augmentation. Methods based on generative models directly translate or generate new samples, whereas methods based on external knowledge utilize triples and their contextual text from external knowledge bases to create samples. Additionally, external knowledge-based methods incorporate adversarial training to enhance the consistency of generated samples with the target domain.} 
    \label{fig: data_aug}
\end{figure}

\subsection{{Data Augmentation}}
{
ECI datasets often have insufficient training samples. Sparse annotations worsen data scarcity, which harms supervised model performance. Data augmentation strategies have been developed for sentence-level causality annotation enhancement. These methods can be categorized into external knowledge-based and generation-based methods (see Figure \ref{fig: data_aug}).}
\subsubsection{{External Knowledge-based Augmentation}}
{
This approach generates training samples using causal relations from external knowledge bases. \textit{KnowDis} \cite{knowdis} uses $\texttt{WordNet}$ and $\texttt{VerbNet}$ \cite{verbnet} to expand causal event pairs, applying a commonsense filter to reduce noise. \textit{LearnDA} \cite{learnda} generates sentences from $\texttt{WordNet}$ and $\texttt{VerbNet}$ causalities and identifies causality to create new training data. \textit{CauSeRL} \cite{causerl} uses self-supervised learning with $\texttt{ATOMIC}$ \cite{atomic} to capture causal patterns, transferring knowledge to ECI models.}
\subsubsection{{Generation-based Augmentation}}
{
This approach generates new samples through techniques like back-translation, prediction, rewriting, and paraphrasing. \textit{ERDAP} \cite{erdap} predicts new event relations using a relational graph convolutional network for augmentation. \textit{PCC} \cite{pcc} uses paraphrasing to create challenging samples, applying cyclic learning to maintain model stability. \textit{EDA} \cite{eda} applies synonym replacement, random insertion, swap, and deletion to generate diverse training examples. \textit{BT-ESupCL} \cite{BT-ESupCL} uses back-translation to create new samples while preserving semantic meaning.}
\subsubsection{{Open Challenges}}
{
Data augmentation improves model generalization but relies on the quality of external knowledge bases and generative models, which may lack domain-specific patterns, causing semantic mismatches. Over-augmentation can introduce noise through incorrect causal relations or distorted samples. Domain adaptation remains challenging, as general-purpose augmented data may not suit specialized domains like clinical or financial ECI.}

\section{Document-Level Event Causality Identification}\label{sec: deci}
Document-level causal links span multiple sentences or paragraphs, requiring models to integrate broader context, resolve coreferences, and handle implicit causality. This complexity limits the effectiveness of traditional syntax and pattern-based methods, necessitating advanced semantic understanding and reasoning approaches. {The initial research on DECI emerged in 2018, with Gao et al. \cite{ilp} introducing a pioneering approach to document-level structure integration. Their methodology focused on identifying "main events" with a high likelihood of engaging in multiple causal relationships, while incorporating syntactic, discourse, and coreference constraints within an ILP framework to optimize causal link predictions across event pairs. However, it wasn't until 2022 that DECI research began gaining substantial momentum in the academic community. Current DECI methods primarily include deep semantic encoding-based models, event graph-based models, and prompt-based fine-tuning models. }

\subsection{{Deep Semantic Encoding}}
{
This approach shares similar principles with the deep semantic encoding approaches previously introduced in the SECI models. However, given their involvement with more complex contextual information, they typically require targeted strategies or enhanced encoding techniques to improve representation quality.
\textit{KADE} \cite{kade} enhances event representations by retrieving related knowledge from $\texttt{ConceptNet}$ and merging it with the original event sentence. For rare events, \textit{KADE} introduces an analogy mechanism, using k-Nearest-Neighbors (kNN) to retrieve similar events from memory and refine representation.
\textit{SENDIR} \cite{sendir} applies sparse attention to selectively focus on relevant information within long texts, differentiating intra- and inter-sentence causality. It then builds reasoning chains for cross-sentence events by identifying supporting events through gated attention.
\textit{DiffusECI} \cite{diffuseci} iteratively adds and removing noise to transform event context representations into refined causal labels. During inference, the model progressively filters noise to enhance causal signal clarity, improving causality predictions across lengthy texts.
{
\textit{Ensemble Learning} \cite{ensemble} combines Mamba \cite{gu2024mamba}, Temporal Convolutional Network (TCN) \cite{tcn}, and GNN to capture causal relationships effectively. It uses DistilBERT \cite{distilbert} for text encoding and dual graphs to enhance local information capture, addressing data imbalance through optimized sampling and weighting.}
}

{
\textit{Summary}. Compared to SECI models, deep semantic encoding-based DECI models face greater challenges including long-context processing, long-range dependency decay, and textual noise interference. Moreover, many lengthy documents exceed the maximum length capacity of encoders, necessitating segmentation approaches that significantly adversely affect model performance. A critical challenge for such methods lies in enhancing document-level event embedding quality—reducing noise while effectively capturing causal-relevant semantic information.
}

\subsection{Event Graph Reasoning-Based Methods}
\begin{figure}[t]
    \centering
    \includegraphics[width=0.95\columnwidth]{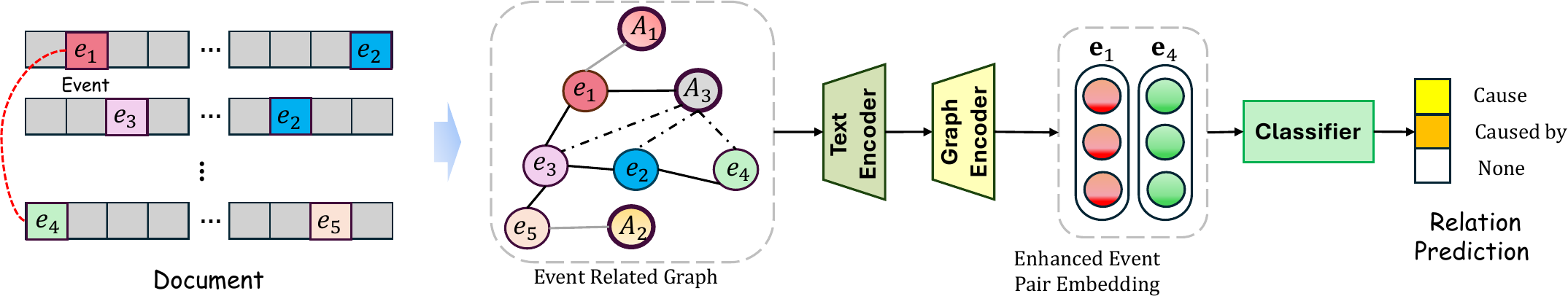}
    \caption{Framework of event graph reasoning for DECI. Event graphs can be homogeneous or heterogeneous, representing various information types—event mentions, event pairs, documents, sentences, and words—as nodes connected by different predefined edges. Graph embedding models encode these nodes, applying diverse strategies to create embeddings for events and event pairs, which are then used by a classifier to predict event causalities. $A_1-A_3$ represents supplementary feature nodes (e.g. context, arguments).} 
    \label{fig: event_graph}
\end{figure}

Event graph reasoning-based DECI methods model document-level causalities by constructing diverse event-related graph structures, framing causality as a reasoning task between nodes and edges (see Figure \ref{fig: event_graph}). Typically, these methods first use PLMs to encode events and context, then build a heterogeneous graph centered on events, treating elements like events, words, and phrases as nodes. Various types of edges (e.g., event co-occurrence, syntactic dependencies, and semantic similarities) capture relationships between nodes. Graph embedding models then aggregate node information, updating event representations to capture global context and infer causalities through high-order or path-based reasoning.

\textit{DCGIM} \cite{dcgim} models events as nodes and causal links as edges, using attention to update sentence representations with document-level context. A directed GCN aggregates node information, and a biaffine layer predicts edge types. \textit{RichGCN} \cite{richgcn} builds a document-level graph with nodes for words, events, and entities, and edges for discourse, syntactic, and semantic relations. A GCN with sparse regularization aggregates information, and an MLP classifies causalities. \textit{GESI} \cite{gesi} constructs a heterogeneous graph with nodes for event mentions, events, and documents, using GCNs and attention to weigh mentions and infer causalities via path-based reasoning. \textit{ERGO} \cite{ergo} creates a high-order event relational graph with event pair nodes and causal link edges, using a Graph Transformer as encoder and an adaptive focal loss to reduce errors. \textit{CHEER} \cite{cheer} builds a high-order event interaction graph, leveraging centrality and multi-layer GCNs with self-attention to adjust edge weights. \textit{DocECI} \cite{doceci} uses a text structure graph (words, sentences) and a mention relation graph (event mentions, sentences), with Relational GCN and multi-head attention to integrate causal information. {\textit{EHNEM} \cite{ehnem} models causal structures via a hypergraph, combining RoBERTa and hypergraph convolutional networks to identify multiple causal relationships.} \textit{PPAT} \cite{ppat} constructs a sentence-boundary event relationship graph, using pairwise attention for local and progressive inter-sentence causality inference. \textit{iLIF} \cite{ilif} builds an event causal graph, iteratively updating event representation and event causality graph based on high-confidence intra-sentence causal links to inform inter-sentence reasoning.

We present the types of nodes and edges used in the graphs of the above models in Tabel \ref{tab:graphintro}.

\begin{table}[htbp]
  \centering
  \caption{The node and edge types in event graph reasoning-based DECI models.}
  \begin{small}
    \begin{tabular}{|c|c|c|}
    \hline
    Model & Nodes & Edges \\
    \hline
    DCGIM \cite{dcgim} & Events, Context & Causalities \\
    \hline
    RichGCN \cite{richgcn} & \makecell{Words, Event mentions, Entity mentions} & \makecell{Discourse association, Syntactic dependency,\\ Semantic similarity} \\
    \hline
    GESI \cite{gesi}  & \makecell{Event mentions, Events, Document} & \makecell{Mention-Mention, Mention-Event, \\ Event-Event, Event-Document} \\
    \hline
    ERGO \cite{ergo}  & Event Pairs & Shared event mentions \\
    \hline
    CHEER \cite{cheer} & Events, Event Pairs & \makecell{Event pair-Event pair, Event-Event, Event pair-Event} \\
    \hline
    DocECI \cite{doceci} & \makecell{Words, Sentences, Event mentions} & \makecell{Word adjacency, Word dependency, \\ Word-Sentence, Sentence adjacency, Sentence supplement} \\
    \hline
    PPAT \cite{ppat} & Event Pairs & Shared event mentions\\
    \hline
    EHNEM \cite{ehnem} & Event Mentions & Hyperedges (connect multiple events) \\
    \hline
    iLIF \cite{ilif} & Event mentions & Intra-sentence causality, Inter-sentence causality \\
    \hline
    %
    \end{tabular}%
 \end{small}
  \label{tab:graphintro}%
\end{table}%

\textit{Summary}. Event graph-based ECI methods effectively model long-distance dependencies by aggregating information across sentences and paragraphs, updating event representations with a global perspective to reduce the impact of contextual noise on causality identification. However, as document size grows with more events, phrases, and elements, the graph size increases, potentially introducing unnecessary edges that add graph noise and affect model accuracy. Additionally, diverse node and edge types require specialized aggregation and reasoning mechanisms, complicating model structure and making practical tuning more challenging.

\subsection{Prompt-based Fine-Tuning}
Seem to models for SECI, prompt-based fine-tuning methods DECI methods use PLMs to convert ECI tasks into generation or mask-filling tasks through carefully designed natural language prompts. By providing clearly defined prompts, these methods dynamically capture context, guiding the model to predict or generate causal links between events.

\textit{CPATT} \cite{cpatt} treats ECI as a natural language generation task. By applying a constrained prefix attention mechanism, it reduces co-occurrence interference and prompt ambiguity. Specific templates for events and relations are used to dynamically generate context-relevant prompts, enhancing the capture of semantic features across relationship types. 
\textit{DAPrompt} \cite{daprompt} bypasses the limitations of manually designed templates by assuming a causal link between events, then testing this through plausibility evaluation. The prompt includes event placeholders and causal assessment masks, with two MLM classifiers calculating the probability of each event matching its placeholder. 
\textit{HOTECI} \cite{hoteci} leverages optimal transport at sentence and word levels to select key context relevant to causality. Using a prompt to ask if a causality exists, the model outputs essential contextual words and a causality label (“yes” or “no”). Reinforcement learning further optimizes context selection by rewarding choices.

\textit{Summary}.  Prompt-based methods leverage extensive knowledge in PLMs and flexible prompt design to adapt to diverse expressions of event relations, offering strong generalization. However, compared to event graph-based methods, prompt-based approaches are better suited for short-distance causal links. When causalities span multiple sentences or paragraphs, these models may struggle with long-distance dependencies due to contextual noise interference.

\subsection{Open Challenges}
DECI research is still in its early stages, and the methods discussed above are not yet part of a cohesive methodological framework. Key challenges in DECI include noise reduction in long texts, high-quality event representation, and event causal chain reasoning. With the development of LLMs, future research will likely focus on refining these aspects to improve DECI performance.

\section{{Emerging Paradigms in ECI}}\label{generalizable}

With the continuous advancement of ECI, recent years have witnessed a paradigm shift from traditional supervised, monolingual approaches to unsupervised and multi-lingual/cross-lingual methodologies. Notably, the development of cutting-edge LLMs and their associated reasoning algorithms has significantly propelled progress in ECI studies. This section will focus on advanced LLM-based ECI approaches and multi-lingual/cross-lingual ECI research.\footnote{Since both of these emerging ECI paradigms require models to possess DECI capabilities, they are also categorized within the scope of DECI research in Table \ref{tab:taxonomy}}.

\subsection{{Large Language Models for ECI}}
{
LLMs can function as CKPLMs due to their extensive training on diverse corpora rich in causal information. As early as 2023, when LLMs gained widespread popularity, Gao et al. \cite{llm-eci} conducted comprehensive research on the performance of ChatGPT-series LLMs in ECI tasks. They employed Chain-of-Thought (CoT) \cite{cot} prompting to facilitate step-by-step reasoning and ICL to provide examples that guide output. However, their experiments revealed that LLMs struggle with causal reasoning. Notably, CoT prompts and example provision can sometimes hinder LLMs' performance in ECI tasks. This is attributed to LLMs' inherent “causal hallucination,” which leads to over-attribution of causality. Moreover, the ECI performance of LLMs is highly sensitive to prompt design, resulting in inconsistent outcomes.
To mitigate the negative effects of causal hallucination in LLMs, researchers have explored various strategies, such as external knowledge constraints, consistency checks, multi-aspects reasoning, and using LLMs as auxiliary tools.}

{
Knowledge-enhanced LLM for ECI improves causal reasoning by integrating structured knowledge and LLMs. \textit{KLop} \cite{KLop} aligns Causal Knowledge Graphs (CKGs) with LLMs using a descriptor module for alignment data and an aligner to bridge modality gaps. \textit{LKCER} \cite{lkcer} builds a concept-level event heterogeneous graph, generates enriched knowledge via LLMs, and uses joint prompt learning to integrate explicit and implicit causalities. \textit{KnowQA} \cite{knowqa} frames ECI as binary question answering, leveraging document-level event structures with single- and multi-turn causal QA strategies. Zhang et al. \cite{zhang_llm} transform ECI into multiple-choice questions, integrating LLM-generated rationales and linearized event causal graphs, trained via multi-task learning (Q→A, Q→R, Q→G). Wang et al. \cite{wang-etal-2024-event-causality} use LLMs for anonymizing event sequences, retrieving documents, generating summaries, and assessing event similarities. Causal effects are estimated with synthetic control units\footnote{This approach focuses on ECI tasks where data is structured as event sequences.}. \textit{Dr. ECI} \cite{dreci} enhances zero-shot ECI by decomposing tasks into sub-tasks, using multi-agent collaboration (Causal Explorer, Mediator Detector, Direct and Indirect Reasoners) to identify implicit and indirect causal relationships with causal structure knowledge.
}

{
\textit{Summary}. Existing research has demonstrated the immense potential of LLMs, and LLM-based ECI is poised to remain a research hotspot in the near future. Mitigating causal hallucination continues to be a significant challenge. The key to addressing this issue lies in enhancing LLMs' ability to distinguish between causality, correlation, temporal relations, and other similar semantics, as well as equipping LLMs with a “causal chain of thought” to enable genuine causal logical reasoning. Furthermore, approaches grounded in causal evidence consistency may yield promising results.
}

\subsection{{Multi-Lingual and Cross-Lingual ECI}}
{
Although ECI in English texts has been extensively studied, research on ECI in other languages remains relatively limited. While existing models can be adapted to different languages by modifying encoders, and texts can be translated to enable seamless model application, variations in syntactic patterns across languages may impact model performance. Consequently, developing multi-lingual ECI models and achieving cross-lingual generalization have garnered significant attention from researchers. Currently, transfer learning methods and LLMs are the primary tools for multi-lingual and cross-lingual ECI research.}

{
{Cross-lingual ECI research} has grown with the MECI dataset \cite{meci}. Zhu et al. \cite{zhu2023chinese} proposed a cross-lingual ECI method using syntactic graph convolution and cross-attention, mapping word vectors into a shared semantic space with cross-lingual embeddings to enhance causality identification. \textit{GIMC} \cite{gimc} supports zero-shot cross-lingual DECI, constructing a heterogeneous graph with event pairs, sentences, phrases, and statements, using GAT and multi-granularity contrastive learning to align causal representations across languages. \textit{PTEKC} \cite{ptekc} integrates multi-lingual event knowledge from $\texttt{ConceptNet}$ into PLMs via a parameter-sharing adapter and pre-training tasks (event masking and link prediction). \textit{Meta-MK} \cite{metamk} combines multi-lingual knowledge from $\texttt{ConceptNet}$ with meta-learning, using ProtoMAML \cite{protomaml} and L2 normalization to extract language-independent causal features.
}

{
\textit{Summary}. Cross-lingual and multi-lingual ECI fundamentally involves transferring models across different semantic contexts and syntactic patterns. However, due to significant syntactic differences in some languages, achieving semantic transfer using models with limited-scale parameters pre-trained on restricted corpora is challenging. Therefore, training multi-lingual LLMs through semantic alignment is likely the most effective approach to address this task.}

\begin{table}[htbp]
\centering
\caption{{Assessment on five aspects of SECI models.\label{tab:seci_assess}}}
{\footnotesize
\begin{tabular}{|c|c|c|c|c|c|c|c|c|c|c|}
\hline
Task & \multicolumn{2}{c|}{Taxonomy} & Model & Conference or \textit{Journal} & Year & I & A & G & E & F \\
\hline
\multirow{49}[18]{*}{SECI} & \multicolumn{2}{c|}{\multirow{3}[1]{*}{\makecell{Feature Pattern-\\based Matching}}} & Template Matching & \textbf{-} & \textbf{-} & × & $\star$ & × & \textbf{-} & $\star$ \\
          & \multicolumn{2}{c|}{} & Syntactic Pattern & \textbf{-} & \textbf{-} & × & $\star$ & × & \textbf{-} & $\star$ \\
          & \multicolumn{2}{c|}{} & Co-occurrence Pattern & \textbf{-} & \textbf{-} & $\sqrt{}$ & × & × & \textbf{-} & × \\
\cline{2-11}
          & \multicolumn{2}{c|}{\multirow{2}[1]{*}{\makecell{Machine Learning-\\based Classification}}} & Explicit Lexical-based & \textbf{-} & \textbf{-} & × & × & × & \textbf{-} & $\sqrt{}$ \\
          & \multicolumn{2}{c|}{} & Implicit Feature-based & \textbf{-} & \textbf{-} & $\sqrt{}$ & × & × & \textbf{-} & × \\
\cline{2-11}
          & \multicolumn{1}{c|}{\multirow{28}[8]{*}{\makecell{Deep Semantic\\Encoding}}} & Simple Encoding & - & - & - & $\sqrt{}$ & × & × & \textbf{-} & $\sqrt{}$ \\
\cline{3-11}
          & & \multicolumn{1}{c|}{\multirow{13}[2]{*}{\makecell{Textual Information\\Enhanced Encoding}}} & Tri-CNN \cite{tri-cnn} & ACL & 2016 & $\sqrt{}$ & × & $\sqrt{}$ & \textbf{-} & $\sqrt{}$ \\
          & & & Ayyanar et al. \cite{ayyanar} & INDICON & 2016 & $\sqrt{}$ & × & $\sqrt{}$ & \textbf{-} & $\sqrt{}$ \\
          & & & Ponti and Korhonen \cite{ponti-korhonen-2017-event} & LSDSem & 2017 & $\sqrt{}$ & × & $\sqrt{}$ & \textbf{-} & $\sqrt{}$ \\
          & & & BiLSTM \cite{bilstm} & EMNLP & 2017 & $\sqrt{}$ & × & $\sqrt{}$ & \textbf{-} & × \\
          & & & Siamese-BiLSTM \cite{siamese} & ICONIP & 2018 & $\sqrt{}$ & × & $\sqrt{}$ & \textbf{-} & $\sqrt{}$ \\
          & & & BERT+MA \cite{bert-multiano} & EMNLP & 2019 & $\sqrt{}$ & × & $\sqrt{}$ & \textbf{-} & $\sqrt{}$ \\
          & & & CEPN \cite{cepn} & WWW & 2022 & $\sqrt{}$ & × & × & \textbf{-} & $\star$ \\
          & & & RPA-GCN \cite{rpa-gcn} & \textit{Information} & 2022 & $\sqrt{}$ & × & $\sqrt{}$ & \textbf{-} & $\sqrt{}$ \\
          & & & SemSIn \cite{semsin} & ACL & 2023 & $\star$ & $\sqrt{}$ & $\star$ & \textbf{-} & $\sqrt{}$ \\
          & & & SCL \cite{scl} & CCIS & 2023 & $\sqrt{}$ & $\sqrt{}$ & $\sqrt{}$ & \textbf{-} & $\sqrt{}$ \\
          & & & ECLEP \cite{eclep} & ACL & 2023 & $\sqrt{}$ & $\sqrt{}$ & $\sqrt{}$ & \textbf{-} & $\sqrt{}$ \\
          & & & CF-ECI \cite{cfeci}& ACL & 2023 & $\sqrt{}$ & $\sqrt{}$ & $\sqrt{}$ & \textbf{-} & $\sqrt{}$ \\
          & & & SemDI \cite{semdi} & EMNLP & 2024 & $\star$ & $\sqrt{}$ & $\star$ & - & $\star$ \\
\cline{3-11}
          & & \multicolumn{1}{c|}{\multirow{14}[4]{*}{\makecell{External Knowledge\\Enhanced Encoding}}} & KnowMMR \cite{knowmmr} & IJCAI & 2020 & $\sqrt{}$ & $\sqrt{}$ & $\sqrt{}$ & $\rightarrow$ & × \\
          & & & DSET \cite{dset} & DOCS & 2023 & $\sqrt{}$ & $\sqrt{}$ & $\star$ & $\rightarrow$ & $\sqrt{}$ \\
          & & & ECIFF \cite{eciff} & ICTAI & 2023 & $\sqrt{}$ & $\star$ & $\sqrt{}$ & $\uparrow$ & $\sqrt{}$ \\
          & & & ECIHS \cite{ecihs} & INTERSPEECH & 2023 & $\sqrt{}$ & $\star$ & $\sqrt{}$ & $\uparrow$ & $\sqrt{}$ \\
          & & & KLPWE \cite{klpwe} & \textit{Machine Learning} & 2024 & $\sqrt{}$ & $\star$ & $\sqrt{}$ & $\uparrow$ & $\sqrt{}$ \\
          & & & DFP \cite{dfp} & COLING & 2024 & $\star$ & $\star$ & $\star$ & $\uparrow$ & $\sqrt{}$ \\
          & & & KIGP \cite{kigp} & \textit{Applied Intelligence} & 2025 & $\sqrt{}$ & $\sqrt{}$ & $\sqrt{}$ & $\rightarrow$ & $\sqrt{}$ \\
\cline{4-11}
          & & & K-CNN \cite{k-cnn} & \textit{ESWA} & 2019 & $\sqrt{}$ & $\sqrt{}$ & $\sqrt{}$ & \textbf{-} & $\sqrt{}$ \\
          & & & LSIN \cite{lsin}& ACL & 2021 & $\sqrt{}$ & $\sqrt{}$ & $\star$ & $\rightarrow$ & $\sqrt{}$ \\
          & & & BERT+MFN \cite{bert+mfn} & \textit{KSII} & 2022 & $\sqrt{}$ & $\sqrt{}$ & $\sqrt{}$ & $\rightarrow$ & $\sqrt{}$ \\
          & & & Chen and Mao \cite{chenandmao} & \textit{ESWA} & 2024 & $\star$ & $\star$ & $\star$ & $\uparrow$ & $\sqrt{}$ \\
          & & & GCKAN \cite{gckan}& \textit{IS} & 2024 & $\sqrt{}$ & $\star$ & $\sqrt{}$ & $\uparrow$ & $\sqrt{}$ \\
          & & & C3NET \cite{c3net} & \textit{KBS} & 2024 & $\star$ & $\sqrt{}$ & $\star$ & $\rightarrow$ & $\sqrt{}$ \\
\cline{2-11}
          & \multicolumn{2}{c|}{\multirow{3}[2]{*}{Causal Knowledge\newline Pre-training}} & Unified-QA \cite{UnifiedQA} & EMNLP & 2020 & $\sqrt{}$ & $\star$ & $\star$ & - & × \\
          & \multicolumn{2}{c|}{} & CausalBERT \cite{causalbert} & \textit{arXiv Preprint} & 2021 & $\sqrt{}$ & $\star$ & $\star$ & \textbf{-} & × \\
          & \multicolumn{2}{c|}{} & Causal-BERT \cite{causal-bert} & \textit{arXiv Preprint} & 2021 & $\sqrt{}$ & $\star$ & $\star$ & \textbf{-} & × \\
\cline{2-11}
          & \multicolumn{2}{c|}{\multirow{8}[2]{*}{Prompt-based Fine-Tuning}} & DPJL \cite{dpjl} & COLING & 2022 & $\sqrt{}$ & $\sqrt{}$ & $\sqrt{}$ & \textbf{-} & $\sqrt{}$ \\
          & \multicolumn{2}{c|}{} & GENECI \cite{geneci} & ACL-SIGLEX & 2022 & $\sqrt{}$ & $\sqrt{}$ & $\sqrt{}$ & \textbf{-} & $\sqrt{}$ \\
          & \multicolumn{2}{c|}{} & KEPT \cite{kept} & \textit{KBS} & 2023 & $\sqrt{}$ & $\star$ & $\star$ & $\uparrow$ & $\sqrt{}$ \\
          & \multicolumn{2}{c|}{} & MRPO \cite{mrpo} & IJCNN & 2023 & $\sqrt{}$ & $\sqrt{}$ & $\sqrt{}$ & \textbf{-} & $\sqrt{}$ \\
          & \multicolumn{2}{c|}{} & RE-CECI \cite{receci} & NLPCC & 2023 & $\sqrt{}$ & $\star$ & $\sqrt{}$ & \textbf{-} & $\sqrt{}$ \\
          & \multicolumn{2}{c|}{} & HFEPA \cite{hfepa} & \textit{Scientific Reports} & 2024 & $\sqrt{}$ & $\sqrt{}$ & $\sqrt{}$ & \textbf{-} & $\sqrt{}$ \\
          & \multicolumn{2}{c|}{} & GenSORL \cite{gensorl} & \textit{KBS} & 2024 & $\sqrt{}$ & $\sqrt{}$ & $\star$ & \textbf{-} & $\sqrt{}$ \\
          & \multicolumn{2}{c|}{} & ICCL \cite{iccl} & EMNLP & 2024 & $\sqrt{}$ & $\star$ & $\sqrt{}$ & - & $\sqrt{}$ \\
\cline{2-11}
          & \multicolumn{1}{c|}{\multirow{5}[3]{*}{Data Augmentation}} & \multicolumn{1}{c|}{\multirow{3}[2]{*}{External Knowledge-based}} & Knowdis \cite{knowdis} & COLING & 2020 & $\sqrt{}$ & $\star$ & $\sqrt{}$ & $\uparrow$ & × \\
          & & & CauSeRL \cite{causerl} & ACL-IJCNLP & 2021 & $\sqrt{}$ & $\star$ & $\sqrt{}$ & $\uparrow$ & × \\
          & & & LearnDA \cite{learnda} & ACL-IJCNLP & 2021 & $\sqrt{}$ & $\star$ & $\sqrt{}$ & $\uparrow$ & × \\
\cline{3-11}
          & & \multicolumn{1}{c|}{\multirow{2}[1]{*}{Generative Model-based}} & ERDAP \cite{erdap} & IAJIT & 2024 & $\sqrt{}$ & $\star$ & $\sqrt{}$ & \textbf{-} & $\sqrt{}$ \\
          & & & BT-ESupCL \cite{BT-ESupCL} & \textit{Neurocomputing} & 2025 & $\sqrt{}$ & $\star$ & $\sqrt{}$ & \textbf{-} & $\sqrt{}$ \\
\hline
\end{tabular}
}
\label{tab:quant_assess1}
\end{table}

\begin{table}[htbp]
\centering
\caption{{Assessment on DECI, LLM for ECI, and Multi-Lingual/Cross-Lingual ECI models.\label{tab:deci_llm_multi_assess}}}
{\footnotesize
\begin{tabular}{|c|c|c|c|c|c|c|c|c|c|c|}
\hline
Task & \multicolumn{2}{c|}{Taxonomy} & Model & Conference or Journal & Year & I & A & G & E & F \\
\hline
\multirow{17}[7]{*}{DECI} & \multicolumn{2}{c|}{\multirow{4}[1]{*}{Deep Semantic Encoding}} & DPLSTM \cite{dplstm} & ACL & 2017 & $\sqrt{}$ & × & $\sqrt{}$ & \textbf{-} & $\sqrt{}$ \\
          & \multicolumn{2}{c|}{} & KADE \cite{kade} & AAAI & 2023 & $\star$ & $\sqrt{}$ & $\star$ & $\uparrow$ & $\sqrt{}$ \\
          & \multicolumn{2}{c|}{} & SENDIR \cite{sendir} & ACL & 2023 & $\star$ & $\sqrt{}$ & $\sqrt{}$ & - & $\sqrt{}$ \\
          & \multicolumn{2}{c|}{} & DiffusECI \cite{diffuseci} & AAAI & 2024 & $\star$ & $\sqrt{}$ & $\star$ & - & $\star$ \\
          & \multicolumn{2}{c|}{} & Ensemble \cite{ensemble} & \textit{Information} & 2025 & $\sqrt{}$ & $\sqrt{}$ & $\sqrt{}$ & - & $\sqrt{}$ \\
\cline{2-11}
          & \multicolumn{2}{c|}{\multirow{9}[2]{*}{Event Graph Reasoning}} & DCGIM \cite{dcgim} & IS & 2021 & $\sqrt{}$ & $\sqrt{}$ & $\sqrt{}$ & - & $\sqrt{}$ \\
          & \multicolumn{2}{c|}{} & RichGCN \cite{richgcn} & NAACL & 2021 & $\sqrt{}$ & $\sqrt{}$ & $\sqrt{}$ & - & × \\
          & \multicolumn{2}{c|}{} & DocECI \cite{doceci} & APWeb-WAIM & 2022 & $\sqrt{}$ & $\sqrt{}$ & $\sqrt{}$ & - & × \\
          & \multicolumn{2}{c|}{} & ERGO \cite{ergo} & COLING & 2022 & $\star$ & $\sqrt{}$ & $\star$ & - & $\sqrt{}$ \\
          & \multicolumn{2}{c|}{} & GESI \cite{gesi} & SIGIR & 2022 & $\sqrt{}$ & $\sqrt{}$ & $\sqrt{}$ & - & $\sqrt{}$ \\
          & \multicolumn{2}{c|}{} & CHEER \cite{cheer} & ACL & 2023 & $\star$ & $\sqrt{}$ & $\star$ & - & $\sqrt{}$ \\
          & \multicolumn{2}{c|}{} & PPAT \cite{ppat} & IJCAI & 2023 & $\star$ & $\sqrt{}$ & $\star$ & - & $\sqrt{}$ \\
          & \multicolumn{2}{c|}{} & iLIF \cite{ilif} & ACL & 2024 & $\star$ & $\sqrt{}$ & $\star$ & - & $\star$ \\
          & \multicolumn{2}{c|}{} & EHNEM \cite{ehnem}& \textit{CSL} & 2025 & $\sqrt{}$ & $\sqrt{}$ & $\star$ & - & $\sqrt{}$ \\
\cline{2-11}
          & \multicolumn{2}{c|}{\multirow{3}[2]{*}{Prompt-based Fine-Tuning}} & CPATT \cite{cpatt} & KBS & 2023 & $\star$ & $\sqrt{}$ & $\star$ & - & $\star$ \\
          & \multicolumn{2}{c|}{} & DAPrompt \cite{daprompt} & ACL & 2023 & $\star$ & $\sqrt{}$ & $\star$ & - & $\sqrt{}$ \\
          & \multicolumn{2}{c|}{} & HOTECI \cite{hoteci} & COLING & 2024 & $\star$ & $\sqrt{}$ & $\star$ & - & $\sqrt{}$ \\
    \hline
    \multicolumn{3}{|c|}{\multirow{7}[2]{*}{LLM for ECI}} & LLM-ECI \cite{llm-eci} & EMNLP & 2023  & $\star$     & $\star$     & $\star$     & -     & × \\
    \multicolumn{3}{|c|}{} & KLop \cite{KLop} & IEEE BigData & 2024  & $\star$     & $\star$     & $\sqrt{}$     & ↑     & $\sqrt{}$ \\
    \multicolumn{3}{|c|}{} & KnowQA \cite{knowqa} & EMNLP & 2024  & $\star$     & $\star$     & $\star$     & -     & × \\
    \multicolumn{3}{|c|}{} & Zhang et al. \cite{zhang_llm} & IJCNN & 2024  & $\star$     & $\sqrt{}$     & $\sqrt{}$     & -     & $\sqrt{}$ \\
    \multicolumn{3}{|c|}{} & Wang et al. \cite{wang-etal-2024-event-causality} & EMNLP & 2024  & $\star$     & $\star$     & $\star$     & -     & $\star$ \\
    \multicolumn{3}{|c|}{} & LKCER \cite{lkcer} & COLING & 2025  & $\star$     & $\sqrt{}$     & $\star$     & -     & $\sqrt{}$ \\
    \multicolumn{3}{|c|}{} & Dr. ECI \cite{dreci} & COLING & 2025  & $\star$     & $\star$     & $\star$     & -     & $\star$ \\
    \hline
    \multicolumn{3}{|c|}{\multirow{4}[2]{*}{Multi-Lingual/Cross-Lingual ECI}} & Zhu et al. \cite{zhu2023chinese} & PRCV  & 2023  & $\sqrt{}$     & $\sqrt{}$     & $\star$     & -     & $\sqrt{}$ \\
    \multicolumn{3}{|c|}{} & GIMC \cite{gimc}  & COLING & 2024  & $\sqrt{}$     & $\star$     & $\star$     & -     & $\sqrt{}$ \\
    \multicolumn{3}{|c|}{} & MetaMK \cite{metamk} & GAIIS & 2024  & $\sqrt{}$     & $\sqrt{}$     & $\star$     & -     &  $\sqrt{}$\\
    \multicolumn{3}{|c|}{} & PTEKC \cite{ptekc} & \textit{INT J MACH LEARN CYB} & 2025  & $\sqrt{}$     & $\sqrt{}$     & $\star$     & -     & $\sqrt{}$ \\
    \hline
\end{tabular}
}
\label{tab:quant_assess2}
\end{table}

\section{Assessment}\label{sec: assess}
In this section, we evaluate the performance of existing ECI models through both qualitative and quantitative assessments.
\subsection{Performance Assessment}\label{perf_assess}

We performed a comprehensive analysis and discussion of all the aforementioned ECI models, comparing them across five key aspects: (1) effectiveness in identifying implicit causality, (2) data annotation requirements, (3) cross-domain generalizability, (4) reliance on external knowledge/tools, and (5) issues related to false positives. For early-stage ECI models, due to their similar performance characteristics, we present evaluation results by category rather than individually for each model. The results of this comparison are summarized in Tables \ref{tab:quant_assess1} and \ref{tab:quant_assess2}\footnote{Note that the results presented in the table are derived from a qualitative, relative comparison, based on an analysis of both the model frameworks and their experimental results. These results are intended to serve as a reference only.}. In the table, "\textbf{I}" assesses the ability of a method to identify implicit causalities($\star$: Strong ability $\sqrt{}$: Moderate ability ×: Weak ability). "\textbf{A}" indicates the level of annotated data required for effective performance (×: High requirement $\sqrt{}$: Moderate requirement $\star$: Low requirement). "\textbf{G}" measures the method’s ability to generalize across different topics or domains ($\star$: Strong generalization capability $\sqrt{}$: Moderate capability ×: Weak capability). "\textbf{E}" indicates the degree to which a method relies on external knowledge (↑: High dependency →: Moderate dependency -: No dependency). "\textbf{F}" assesses the extent to which a method incorrectly reports false positives (×: Severe false positives$\sqrt{}$: Minimal false positives $\star$: Little to no false positives).

{\textbf{Summary}. Deep semantic encoding and its' enhancement strategies are the most effective in capturing implicit causalities, but they also face challenges such as a higher risk of generating false positives and greater data requirements. In contrast, data augmentation strategies and LLM-based approaches have relatively lower data demands. External knowledge-enhanced methods exhibit stronger generalization capabilities and reduced data dependency, but their effectiveness depends on the quantity and quality of the external knowledge base. Prompt-based fine-tuning methods demonstrate relatively comprehensive performance, though, as previously mentioned, they struggle with long-text processing. Event graph modeling-based approaches also show balanced performance, but their cross-domain generalization ability needs improvement. LLM-based methods excel in zero-shot settings, yet causal hallucinations lead to significant limitations due to false positives.}

\subsection{Experimental Evaluation}
{
To evaluate the performance of current ECI models, we conducted experiments of several state-of-the-art models using four benchmark datasets: CTB \cite{ctb}, ESL \cite{esc}, MAVEN-ERE \cite{maven}, and MECI \cite{meci}. The first three datasets represent monolingual collections, whereas the final dataset constitutes a multi-lingual corpus. We compared the performance of individual models and conducted a comparative analysis of the performance differences among various categories of methods. Additionally, we delved into the variations in dataset annotation quality and their impact on model performance. \textbf{Due to space limitations, this section is presented in the Appendix \ref{appendix}.}}

\section{Future Directions}\label{sec:challenge}
In the previous sections, we have reviewed the recent advances in ECI. Although more and more models have been introduced to enhance event representation and causal reasoning, there are still many challenges and open problems that need to be addressed.
In this section, we discuss the future directions of this research area.

\subsection{Causal Direction Identification}
Current ECI research primarily focuses on determining the existence of causalities, often overlooking the direction of causality. However, causal direction is crucial for information extraction and knowledge reasoning, especially in constructing a comprehensive event timeline to inform analysis and decision-making \cite{ilif}. Identifying causal direction requires a deeper semantic understanding and representation learning. Future research could explore directed graph-based inference methods, potentially integrating PLMs, GNNs, and causal inference techniques to enhance directional reasoning \cite{causalnlp}.

\subsection{Event Causal Chain Reasoning}
Identified causal relations can be linked to form causal chains, supporting decision-making and providing insights into event development. Understanding these chains is essential for tracing key events and analyzing the sequence of occurrences. However, simply concatenating causal event pairs may be inadequate due to issues like context drift and threshold effects \cite{reco}. Future research should investigate causal chain reasoning, leveraging traditional causal inference methods such as counterfactual data augmentation \cite{counterfact} and Structural Causal Model (SCM) \cite{Pearl2009} to build reliable causal chains.

\subsection{Uncertain ECI}
In real-world scenarios, causality often involves degrees of uncertainty, requiring a "confidence level" or "conditional" reasoning approach. This introduces challenges in implementing conditional and fuzzy reasoning \cite{uncertain} within ECI frameworks. Additionally, existing labeled datasets lack confidence information, typically providing only binary or ternary causal labels. Future efforts should focus on developing datasets that include confidence annotations, despite the complex labeling process, as these are essential for real-world applications.

\subsection{Event Causality Extraction}
ECI typically assumes given event mentions, but identifying relevant events in text first requires event extraction (EE) \cite{ee}. Most studies use EE algorithms to extract events and trigger words before applying ECI, which can introduce error propagation. A more effective approach may involve simultaneous event extraction and causal identification, termed Event Causality Extraction (ECE) or Event Causality Generation (ECG) \cite{seag}. ECE is more complex than ECI but offers significant potential, as event classification and argument extraction in EE help clarify semantic and logical connections between events.


\subsection{Multimodal ECI}
In most ECI tasks, events and contextual information are derived solely from text. However, with advancements in multimodal technologies, events might originate from images or videos, with additional context provided by these sources. Leveraging multimodal data could significantly enhance causal reasoning \cite{multimodal} but requires entity disambiguation across modalities and alignment of semantic representations using multimodal pre-trained models. Developing multimodal ECI datasets is also crucial, potentially achieved by refining existing multimodal relation extraction datasets \cite{redataset1,redataset2}.

\subsection{Explainable ECI}
Explainability is vital in causal inference research. Beyond detecting causalities, understanding the mechanisms and reasoning paths behind them is essential. Current state-of-the-art methods rely heavily on "black-box" deep neural networks, which lack transparency and reduce user trust \cite{xu_interpretability_2025}. Future work should focus on developing methods that provide interpretable causal explanations, clarifying the reasoning steps and evidence supporting causal conclusions. Enhanced explainability will improve model reliability and applicability to real-world tasks \cite{explainable_ai}.

\subsection{Few-Shot and Zero-Shot ECI}
{
While some methods \cite{llm-eci,iccl,knowqa,dreci} exhibit few-shot and zero-shot ECI capabilities, they still face the challenge of causal halluciantion and self inconsistency of LLMs. Future research could address these issues by using LLM-based data augmentation on existing labeled datasets \cite{llm_aug} to generate diverse causal syntactic and grammatical patterns, or by converting text to semantic graph formats to constrain LLM outputs \cite{kg4hallucination}, thereby mitigating causal hallucination problems.}

\subsection{{Integrating Interdisciplinary Perspectives for ECI}}
{
Human causal reasoning often relies on cognitive processes like intuitive judgments, mental models, and contextual biases, which current ECI models do not fully capture. Integrating psychology and cognitive science insights could enhance ECI by aligning models with human-like reasoning. Future research should: (1) incorporate cognitive frameworks to improve causal direction and chain reasoning \cite{gxchen2025languageagentsmirrorhuman}, and (2) use psychological theories of event perception \cite{Zacks2007EventPerception} to ECI. Interdisciplinary datasets combining linguistic, visual, and cognitive annotations could further support models in emulating human causal inference.}

\subsection{{Multi-relation Joint Extraction to Support ECI}}
{
Multi-relation joint extraction enhances ECI by capturing temporal, coreference, and hierarchical links alongside causal relations \cite{cheer,xue-etal-2024-autore}. This approach provides richer context, disambiguates causal links, and improves causal inference robustness across datasets. Joint modeling of multiple relations can also reduce overfitting to specific causal patterns, enhancing model generalizability. However, challenges include diverse feature requirements for different relations and balancing multiple extractors \cite{zhang-etal-2024-srf}. Future work should explore hybrid architectures combining relation-specific and shared models and develop multi-relation datasets like MAVEN-ERE to advance this direction.}

\subsection{{Standard Protocol for Dataset Construction}}
{
The construction of datasets for ECI currently faces challenges due to the subjective nature of causality, leading to inconsistent annotations and varied definitions \cite{cnc}. The lack of a unified causality definition among annotators create non-comparable datasets, complicating model training and evaluation. Future efforts should establish standardized annotation guidelines for explicit and implicit causal relationships and develop a unified causality framework to enhance dataset quality and comparability. Leveraging LLMs for pre-annotation can improve efficiency and scalability \cite{llm_annotate}, but human review is essential to address potential biases and misidentifications, ensuring accuracy through human-AI collaboration.}

\bibliographystyle{ACM-Reference-Format}
\bibliography{main.bib}

\newpage
\appendix
\section{Experimental Evaluation}\label{appendix}
{
To assess the capabilities of existing ECI models, we evaluated multiple state-of-the-art models across four benchmark datasets: CTB \cite{ctb}, ESL \cite{esc}, MAVEN-ERE \cite{maven}, and MECI \cite{meci}. The first three datasets are monolingual, while MECI is a multi-lingual collection. Our analysis included both individual model performance and cross-category comparisons to highlight methodological differences. Furthermore, we examined how discrepancies in annotation quality across datasets influence model effectiveness.}

\subsection{Experimental Setup}
{
\textbf{Dataset Processing}. The details of the four datasets used in this experiment, CTB \cite{ctb}, Event StoryLine v0.9 (ESL) \cite{esc}, MAVEN-ERE \cite{maven}, and MECI \cite{meci} are provided in Section \ref{dataset}. For consistency with prior studies, we adopted the common data splits: in the ESL dataset, the last two topics were designated as development data, while the remaining 20 topics were used for 5-fold cross-validation. For the CTB dataset, we conducted 10-fold cross-validation. For MAVEN-ERE dataset, we leveraged the development set as the test set and split 10\% of the train set for validation.\footnote{Note that Dr. ECI and CPATT employed distinct data processing methods compared to other baselines, as indicated by the superscript delta symbol ($^\Delta$).}}

\textbf{Evaluation Protocol}. We evaluated the models using Precision (P), Recall (R), and F1-score (F1) as metrics, distinguishing between intra-sentence causality, inter-sentence causality, and overall performance. Given that most existing methods evaluate only the presence of causalities without considering directional accuracy, we focused our comparison similarly on causality existence alone. For the CTB dataset, where only 18 inter-sentence causal pairs exist, we limited evaluation to intra-sentence causality. For the ESL and MAVEN-ERE dataset, we assessed performance on intra-sentence causality, inter-sentence causality, and overall identification. For MECI, we evaluated performance of all models on five different language sub-sets.

{
\textbf{Baselines}. We included several state-of-the-art models from some of the existing literature, along with standard baselines like LSTM and BERT. In MECI, we used multi-lingual PLM XLM-R \cite{xlmr} as the backbone. Specifically, we selected those models that have reported results in the literature or have publicly available source code as our baselines. Additionally, we incorporated multiple LLMs as baselines, including LLaMA2-7B\footnote{\url{http://huggingface.co/meta-llama/Llmma-2-7b}}, GPT-3 (text-davinci), GPT-3.5-turbo, GPT-4, GPT-4o-mini\footnote{\url{https://openai.com/chatgpt/}}, DeepSeek-Chat (V3), DeepSeek-R1\footnote{\url{https://www.deepseek.com/}}. For the LLMs, we used two standardized prompt: 
\begin{itemize}
    \item \textit{[D] [SEP] Is there a causal relationship between [E1] and [E2]?}
    \item \textit{[D] [SEP] Does [E1] cause [E2]? }
\end{itemize}
Here, if \textit{[E1]} and \textit{[E2]} are within the same sentence, \textit{D} represents that sentence; otherwise, \textit{D} represents the entire document. It should be noted that many DECI models also support SECI, so models specifically designed for DECI are included in SECI performance comparisons where applicable. We adopt distinct prompting strategies: (1) the first prompt template for GPT-3/GPT-4/LLaMA2-7B to elicit direct generations, and (2) the second template plus bidirectional reasoning validation (E1→E2 and E2→E1 inference) for DeepSeek-Chat/DeepSeek-R1/GPT-4o-mini, where causal relationships are only confirmed when both inference directions yield logically consistent results.

{
\textbf{Implementation Details}. To ensure fair comparisons, if baseline models had reported results on the four datasets, we directly cited those results. For baselines where the original paper used different datasets or splits but provided open-source code, we re-evaluated the model by reproducing it according to our specified data splits. Reproduction strictly followed hyperparameters reported in the original paper or code to maintain result comparability. Models without open-source code or using entirely different datasets were excluded. For models with multiple variants, we reported the variant with the highest F1-score. \textit{In the results table, we mark our reproduced results with an asterisk symbol (*)}. For the CTB and ESL datasets, we strived to reproduce all reproducible results. For the MAVEN-ERE and MECI datasets, given its large scale and complexity, we only reproduced/reported results of the latest methods proposed/published after 2023 (if not reported in the original literature). For simple encoding models, we set \( \mathbf{ep}_{ij} = [\mathbf e_{\text{[CLS]}}; \mathbf{e}_i; \mathbf{e}_j] \) and leveraged equation \ref{simpreason} for the results. }

\begin{table}[htbp]
  \centering
  \caption{{Experimental Results in CTB. All the results have been multiplied by 100, the \textbf{bold} values represent the optimal results, while the \underline{underlined} values indicate the suboptimal results.}}
  \begin{scriptsize}
    \begin{tabular}{|c|c|c|c|c|c|}
       \hline
    \multicolumn{2}{|c|}{\multirow{2}[2]{*}{Type}} & \multirow{2}[2]{*}{Model} & \multicolumn{3}{c|}{Intra-sentence} \\
\cline{4-6}    \multicolumn{2}{|c|}{} &       & P     & R     & F1 \\
    \hline
    \multicolumn{2}{|c|}{\multirow{2}[2]{*}{Feature Pattern -Based Matching}} & RB    & 36.8  & 12.3  & 18.4 \\
    \multicolumn{2}{|c|}{} & DD    & 67.3  & 22.6  & 33.9 \\
\hline    \multicolumn{2}{|c|}{ Feature Classification} & VR-C  & 69.0  & 31.5  & 43.2 \\
    \hline
    \multicolumn{1}{|c|}{\multirow{23}[10]{*}{Deep Semantic Encoding}} & \multirow{5}[2]{*}{Simple Encoding} & LSTM* & 18.3  & 43.5  & 25.8 \\
          & \multicolumn{1}{c|}{} & BiLSTM* & 21.5  & 46.7  & 29.4 \\
          & \multicolumn{1}{c|}{} & BERT  & 41.4  & 45.8  & 43.5 \\
          & \multicolumn{1}{c|}{} & RoBERTa & 39.9  & 60.9  & 48.2 \\
          & \multicolumn{1}{c|}{} & Longformer* & 29.6  & 68.6  & 41.4 \\
\cline{2-6}          & \multirow{4}[2]{*}{\makecell{Textual Information \\ Enhanced Encoding}} & SemSIn & 52.3  & 65.8  & 58.3 \\
          & \multicolumn{1}{c|}{} & ECLEP & 50.6  & 63.4  & 56.3 \\
          & \multicolumn{1}{c|}{} & CF-ECI & 50.5  & 59.9  & 54.8 \\
          & \multicolumn{1}{c|}{} & SemDI & 59.3  & 77.8  & 67.0 \\
\cline{2-6}          & \multirow{10}[4]{*}{\makecell{External Knowledge \\ Enhanced Encoding}} & KnowMMR & 36.6  & 55.6  & 44.1 \\
          & \multicolumn{1}{c|}{} & DSET  & 65.8  & 72.1  & 68.5 \\
          & \multicolumn{1}{c|}{} & ECIFF & 53.1  & 59.4  & 56.1 \\
          & \multicolumn{1}{c|}{} & ECIHS & 61.4  & 64.6  & 63.0 \\
          & \multicolumn{1}{c|}{} & DFP   & 63.7  & 64.2  & 58.5 \\
\cline{3-6}          & \multicolumn{1}{c|}{} & KIGP  & 61.3  & 63.4  & 62.3 \\
          & \multicolumn{1}{c|}{} & C3NET & 60.2  & 72.8  & 65.9 \\
          & \multicolumn{1}{c|}{} & Chen and Mao & \underline{83.7}  & 55.5  & 65.2 \\
          & \multicolumn{1}{c|}{} & GCKAN & 52.2  & 60.7  & 56.1 \\
          & \multicolumn{1}{c|}{} & LSIN  & 51.5  & 56.2  & 52.9 \\
\cline{2-6}          & \multirow{4}[2]{*}{DECI Models} & DiffusECI & \textbf{87.7}  & 66.1  & \underline{75.4} \\
          & \multicolumn{1}{c|}{} & KADE  & 56.8  & 70.6  & 66.7 \\
          & \multicolumn{1}{c|}{} & SENDIR & 65.2  & 57.7  & 61.2 \\
          & \multicolumn{1}{c|}{} & Ensemble & 39.5  & 57.3  & 46.1 \\
    \hline
    \multicolumn{2}{|c|}{\multirow{8}[4]{*}{Prompt-based Fine-Tuning}} & DPJL  & 63.6  & 66.7  & 64.6 \\
    \multicolumn{2}{|c|}{} & GENECI & 60.1  & 53.3  & 56.5 \\
    \multicolumn{2}{|c|}{} & KEPT  & 48.2  & 60    & 53.5 \\
    \multicolumn{2}{|c|}{} & GenSORL & 60.1  & 53.3  & 56.3 \\
    \multicolumn{2}{|c|}{} & ICCL  & 63.7  & 68.8  & 65.4 \\
\cline{3-6}    \multicolumn{2}{|c|}{} & CPATT$^\Delta$ & 77.5  & 73.2  & 75.2 \\
    \multicolumn{2}{|c|}{} & DAPrompt & 66.3  & 67.1  & 65.9 \\
    \multicolumn{2}{|c|}{} & HOTECI & 71.1  & 65.9  & 68.4 \\
    \hline
    \multicolumn{2}{|c|}{\multirow{4}[1]{*}{Data Augmentation}} & CauSeRL & 43.6  & 68.1  & 53.2 \\
    \multicolumn{2}{|c|}{} & Knowdis & 42.3  & 60.5  & 49.8 \\
    \multicolumn{2}{|c|}{} & LearnDA & 41.9  & 68.0  & 51.9 \\
    \multicolumn{2}{|c|}{} & \multicolumn{1}{c|}{BT-ESupCL } & 59.1  & 64.3  & 58.5 \\
    \hline
    \multicolumn{2}{|c|}{\multirow{5}[1]{*}{Event Graph Reasoning}} & CHEER & 56.4  & 69.5  & 62.3 \\
    \multicolumn{2}{|c|}{} & ERGO  & 62.1  & 61.3  & 61.7 \\
    \multicolumn{2}{|c|}{} & iLIF  & 69.9  & 46.7  & 56.0 \\
    \multicolumn{2}{|c|}{} & PPAT  & 67.9  & 64.6  & 66.2 \\
    \multicolumn{2}{|c|}{} & RichGCN & 39.7  & 56.5  & 46.7 \\
    \hline
    \multicolumn{1}{|c|}{\multirow{12}[6]{*}{LLM-based}} & \multirow{5}[2]{*}{\makecell{Simple LLM \\ w/o consistency}} & LLaMA2-7B  & 5.4   & 53.9  & 9.8 \\
          & \multicolumn{1}{c|}{} & text-davinci-002 & 5.0   & 75.2  & 9.3 \\
          & \multicolumn{1}{c|}{} & text-davinci-003 & 8.5   & 64.4  & 15.0 \\
          & \multicolumn{1}{c|}{} & GPT-3.5-turbo & 7.0   & 82.6  & 12.8 \\
          & \multicolumn{1}{c|}{} & GPT-4.0  & 6.1   & \textbf{97.4}  & 11.5 \\
\cline{2-6}          & \multirow{3}[2]{*}{\makecell{Simple LLM \\ w consistency}} & GPT-4o-mini* & 30.8  & 48.3  & 34.4 \\
          & \multicolumn{1}{c|}{} & DeepSeek-Chat* & 37.8  & 53.2  & 40.6 \\
          & \multicolumn{1}{c|}{} & DeepSeek-R1* & 43.7  & 59.2  & 37.8 \\
\cline{2-6}      
          & \multicolumn{2}{c|}{Zhang et al. } & 58.8  & 63.8  & 61.2 \\
          & \multicolumn{2}{c|}{LKCER} & 61.0   & 76.3  & 67.3 \\
          & \multicolumn{2}{c|}{Dr. ECI} & 11.7  & 71.8  & 13.0 \\
          & \multicolumn{2}{c|}{Dr. ECI$^\Delta$} & 80.2  & 84.3  & \textbf{82.2} \\
    \hline
    \end{tabular}%
    \end{scriptsize}
  \label{tab:ctb}%
\end{table}%

\begin{table*}[htbp]
  \centering
  \caption{{Experimental Results in ESL. All the results have been multiplied by 100, the \textbf{bold} values represent the optimal results, while the \underline{underlined} values indicate the suboptimal results.}}
  \begin{scriptsize}
    \begin{tabular}{|c|c|c|c|c|c|c|c|c|c|c|c|}
    \hline
    \multicolumn{2}{|c|}{\multirow{2}[1]{*}{Type}} & \multirow{2}[1]{*}{Model} & \multicolumn{3}{c|}{Intra-sentence} & \multicolumn{3}{c|}{Inter-sentence} \\
\cline{4-9}    \multicolumn{2}{|c|}{} &       & P     & R     & F1    & P     & R     & F1 \\
    \hline
    \multicolumn{2}{|c|}{Feature Pattern-based Matching} & OP    & 22.5  & \textbf{98.6} & 36.6  & 3.0   & 40.7  & 5.5 \\
    \hline
    \multicolumn{2}{|c|}{\multirow{2}[2]{*}{ Feature Classification}} & LIP   & 38.8  & 52.4  & 44.6  & 35.1  & 48.2  & 40.6 \\
    \multicolumn{2}{|c|}{} & LR+   & 37.0  & 45.2  & 40.7  & 25.2  & 48.1  & 33.1 \\
    \hline
    \multicolumn{1}{|c|}{\multirow{23}[10]{*}{Deep Semantic Encoding}} & \multirow{5}[2]{*}{Simple Encoding} & LSTM  & 34.0  & 41.5  & 37.4  & 13.5  & 30.3  & 18.7 \\
          & \multicolumn{1}{c|}{} & BiLSTM & 32.7  & 44.9  & 37.8  & 11.3  & 29.5  & 16.4 \\
          & \multicolumn{1}{c|}{} & BERT  & 60.4  & 45.7  & 52.0  & 30.6  & 39.1  & 34.3 \\
          & \multicolumn{1}{c|}{} & RoBERTa & 62.7  & 45.4  & 52.7  & 32.7  & 38.3  & 35.3 \\
          & \multicolumn{1}{c|}{} & Longformer & 47.7  & 69.3  & 56.5  & 26.1  & 55.6  & 35.5 \\
\cline{2-9}          & \multirow{4}[2]{*}{Enhanced Encoding} & SemSIn & 64.2  & 65.7  & 64.9  & -     & -     & - \\
          & \multicolumn{1}{c|}{} & ECLEP & 49.3  & 68.1  & 57.1  & -     & -     & - \\
          & \multicolumn{1}{c|}{} & CF-ECI & 47.1  & 66.4  & 55.1  & -     & -     & - \\
          & \multicolumn{1}{c|}{} & SemDI & 56.7  & 68.6  & 62.0  & -     & -     & - \\
\cline{2-9}          & \multirow{10}[4]{*}{External Knowledge Enhancement} & KnowMMR & 41.9  & 62.5  & 50.1  & -     & -     & - \\
          & \multicolumn{1}{c|}{} & DSET  & 67.4  & 71.8  & 69.5  & -     & -     & - \\
          & \multicolumn{1}{c|}{} & ECIFF & 51.9  & 62.8  & 56.8  & -     & -     & - \\
          & \multicolumn{1}{c|}{} & ECIHS & -     & -     & 65.2  & -     & -     & - \\
          & \multicolumn{1}{c|}{} & DFP   & 55.9  & 69.8  & 62.1  & -     & -     & - \\
\cline{3-9}          & \multicolumn{1}{c|}{} & KIGP  & 62.4  & 68.5  & 65.3  &       &       &  \\
          & \multicolumn{1}{c|}{} & C3NET & 60.5  & 73.6  & 66.4  & -     & -     & - \\
          & \multicolumn{1}{c|}{} & Chen and Mao & 67.4  & 69.5  & 68.3  & -     & -     & - \\
          & \multicolumn{1}{c|}{} & GCKAN & 77.6  & 50.9  & 60.6  & -     & -     & - \\
          & \multicolumn{1}{c|}{} & LSIN  & 49.7  & 58.1  & 52.5  & -     & -     & - \\
\cline{2-9}          & \multirow{4}[2]{*}{DECI Models} & DiffusECI & 65.3  & 78.3  & 71.4  & 61.9  & 59.9  & 60.9 \\
          & \multicolumn{1}{c|}{} & KADE  & 61.5  & 73.2  & 66.8  & 51.2  & 74.2  & 60.5 \\
          & \multicolumn{1}{c|}{} & SENDIR & 65.8  & 66.7  & 66.2  & 33    & \textbf{90} & 48.3 \\
          & \multicolumn{1}{c|}{} & Ensemble & -     & -     & -     & 40.1  & 49.5  & 43.7 \\
    \hline
    \multicolumn{2}{|c|}{\multirow{9}[4]{*}{Prompt-based Fine-Tuning}} & DPJL  & 65.3  & 70.8  & 67.9  & -     & -     & - \\
    \multicolumn{2}{|c|}{} & GENECI & 59.5  & 57.1  & 58.8  & -     & -     & - \\
    \multicolumn{2}{|c|}{} & KEPT  & 50.0  & 68.8  & 57.9  & -     & -     & - \\
    \multicolumn{2}{|c|}{} & GenSORL & 65.6  & 63.3  & 64.6  & -     & -     & - \\
    \multicolumn{2}{|c|}{} & ICCL  & 62.6  & 66.1  & 64.2  & -     & -     & - \\
\cline{3-9}    \multicolumn{2}{|c|}{} & HFEPA & 52    & 65.5  & 57.9  &       &       &  \\
    \multicolumn{2}{|c|}{} & CPATT $^\Delta$ & \textbf{79.4} & 81.3  & \underline{80.4}  & \underline{74.9}  & 60.1  & \underline{66.7} \\
    \multicolumn{2}{|c|}{} & DAPrompt & 64.5  & 73.6  & 68.5  & 59.9  & 59.3  & 59.0 \\
    \multicolumn{2}{|c|}{} & HOTECI & 66.1  & 72.3  & 69.1  & \textbf{81.4} & 40.6  & 55.1 \\
    \hline
    \multicolumn{2}{|c|}{\multirow{4}[1]{*}{Data Augmentation}} & CauSeRL & 41.9  & 69    & 52.1  & -     & -     & - \\
    \multicolumn{2}{|c|}{} & Knowdis & 39.7  & 66.5  & 49.7  & -     & -     & - \\
    \multicolumn{2}{|c|}{} & LearnDA & 42.2  & 69.8  & 52.6  & -     & -     & - \\
    \multicolumn{2}{|c|}{} & BT-ESupCL & 55.5  & 64.2  & 59.3  &       &       &  \\
    \hline
    \multicolumn{2}{|c|}{\multirow{8}[1]{*}{Event Graph Reasoning}} & CHEER & 56.9  & 69.6  & 62.6  & 45.2  & 52.1  & 48.4 \\
    \multicolumn{2}{|c|}{} & DocECI & 45.8  & 67.9  & 54.7  & 44.3  & 59.5  & 50.8 \\
    \multicolumn{2}{|c|}{} & ERGO  & 57.5  & 72.0  & 63.9  & 51.6  & 43.3  & 47.1 \\
    \multicolumn{2}{|c|}{} & GESI  & -     & -     & 50.3  & -     & -     & 49.3 \\
    \multicolumn{2}{|c|}{} & iLIF  & 76.8  & 66.3  & 71.2  & 53.5  & 65.9  & 59.1 \\
    \multicolumn{2}{|c|}{} & PPAT  & 62.1  & 68.8  & 65.3  & 54.0  & 50.2  & 52.0 \\
    \multicolumn{2}{|c|}{} & RichGCN & 49.2  & 63.0  & 55.2  & 39.2  & 45.7  & 42.2 \\
    \multicolumn{2}{|c|}{} & EHNEM & 63.2  & 70.4  & 66.6  & 62.3  & 59.9  & 61.0 \\
    \hline
    \multicolumn{1}{|c|}{\multirow{12}[6]{*}{LLMs for ECI}} & \multirow{5}[2]{*}{Simple LLM w/o consistency} & LLaMA2-7B* & 12.1  & 50.7  & 19.5  & 6.5   & 51.5  & 11.5 \\
          & \multicolumn{1}{c|}{} & text-davinci-002* & 23.2  & 80.0  & 36.0  & 11.4  & 58.4  & 19.1 \\
          & \multicolumn{1}{c|}{} & text-davinci-003* & 33.2  & 74.4  & 45.9  & 12.7  & 54.3  & 20.6 \\
          & \multicolumn{1}{c|}{} & GPT-3.5-turbo* & 27.6  & 80.2  & 41.0  & 15.2  & 61.8  & 24.4 \\
          & \multicolumn{1}{c|}{} & GPT-4.0* & 27.2  & \underline{94.7}  & 42.2  & 16.9  & 64.7  & 26.8 \\
\cline{2-9}          & \multirow{3}[2]{*}{Simple LLM w consistency} & GPT-4o-mini* & 31.1  & 64.7  & 33.8  & 6.2   & 50    & 10.6 \\
          & \multicolumn{1}{c|}{} & DeepSeek-Chat* & 36.6  & 62.3  & 44.1  & 17.9  & 37.7  & 21.9 \\
          & \multicolumn{1}{c|}{} & DeepSeek-R1* & 22.7  & 90    & 35.1  & 17.2  & 65.8  & 22.4 \\
\cline{2-9}     
          & \multicolumn{2}{c|}{Zhang et al. } & -     & -     & 65.3  & -     & -     & 47.2 \\
          & \multicolumn{2}{c|}{LKCER} & 67.3  & 72.7  & 69.8  & -     & -     & - \\
          & \multicolumn{2}{c|}{Dr. ECI} & {29.0}  & 75.4  & 41.9 & -  & -  & - \\
          & \multicolumn{2}{c|}{Dr. ECI$^\Delta$} & \underline{77.6}  & 91.7  & \textbf{84.1} & 68.4  & \underline{86.6}  & \textbf{76.4} \\
    \hline
    \end{tabular}%
  \end{scriptsize}
  \label{tab:ESL}%
\end{table*}%

\subsection{Results and Analysis}
{
Tables \ref{tab:ctb}, \ref{tab:ESL}, \ref{tab:maven}, and \ref{tab:meci} present the results of baseline methods on the four experimental datasets, respectively. }

\textbf{Overall performance}.
On the CTB dataset, \textit{Dr. ECI$^{\Delta}$} achieved the highest F1 score, followed by \textit{DiffusECI}, which had the highest precision, with Chen and Mao \cite{chenandmao} ranking second. \textit{GPT-4.0} led in recall, followed by \textit{Dr. ECI$^{\Delta}$}. On the ESL dataset, \textit{Dr. ECI$^{\Delta}$} outperformed others in intra-sentence causality F1, with \textit{CPATT$^{\Delta}$} second and highest in precision, while \textit{OP} led in recall, followed by \textit{GPT-4.0}. For inter-sentence causality, \textit{Dr. ECI$^{\Delta}$} again topped F1, with \textit{CPATT$^{\Delta}$} second, \textit{HOTECI} leading in precision, and \textit{SENDIR} in recall. On the MAVEN-ERE dataset, \textit{Dr. ECI$^{\Delta}$} excelled in both intra- and inter-sentence causality F1 and precision, with \textit{ERGO} and \textit{iLIF} as notable runners-up, and \textit{Dr. ECI$^{\Delta}$} also led in recall, followed by \textit{GPT-4.0} and \textit{ERGO}. On the MECI dataset, \textit{DiffusECI} and \textit{HOTECI} led in English, \textit{CHEER} in Danish, \textit{GIMC} in Spanish, and \textit{HOTECI} in Turkish and Urdu, with \textit{GIMC}, \textit{CHEER}, and \textit{Meta-MK} as strong contenders. Overall, \textit{Dr. ECI$^{\Delta}$}, \textit{CPATT$^{\Delta}$}, \textit{HOTECI}, and \textit{GIMC} consistently showed robust performance across datasets and languages.

\begin{figure}[t]
    \centering
    \subfloat[CTB (Intra-Sentence)]{%
        \includegraphics[width=0.95\textwidth]{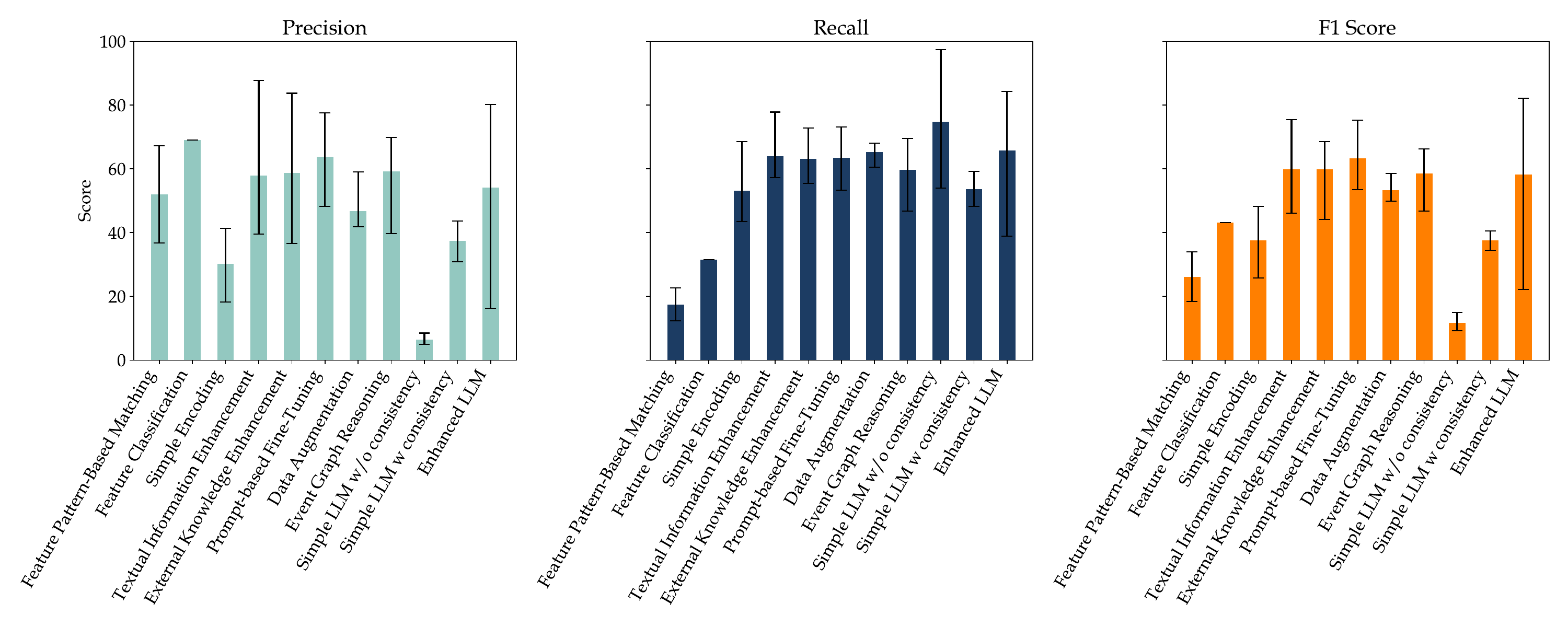}
        \label{fig:ctb_intra}
    }
    \caption{Performance comparison (Precision, Recall, F1) on CTB dataset. Error bars represent the range of variability across models.}
    \label{fig:ctb}
\end{figure}

To analyze the performance of different types of methods, we calculated the average performance for each method category. Figures \ref{fig:ctb} to \ref{fig:meci} display the comparative results across these method types.

\begin{figure}[htbp]
    \centering
    \subfloat[ESL (Intra-Sentence)]{%
        \includegraphics[width=0.95\textwidth]{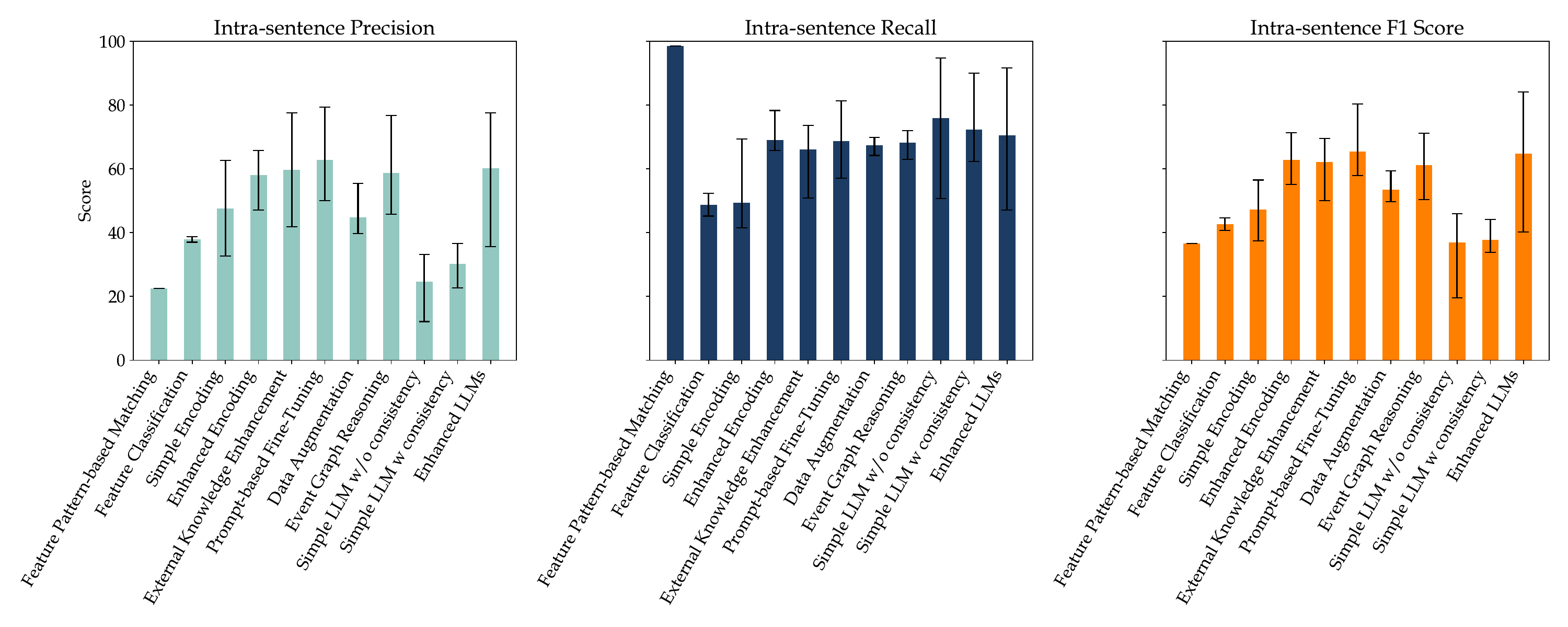}
        \label{fig:esl_intra}
    }\hfill
    \subfloat[ESL (Inter-Sentence)]{%
        \includegraphics[width=0.95\textwidth]{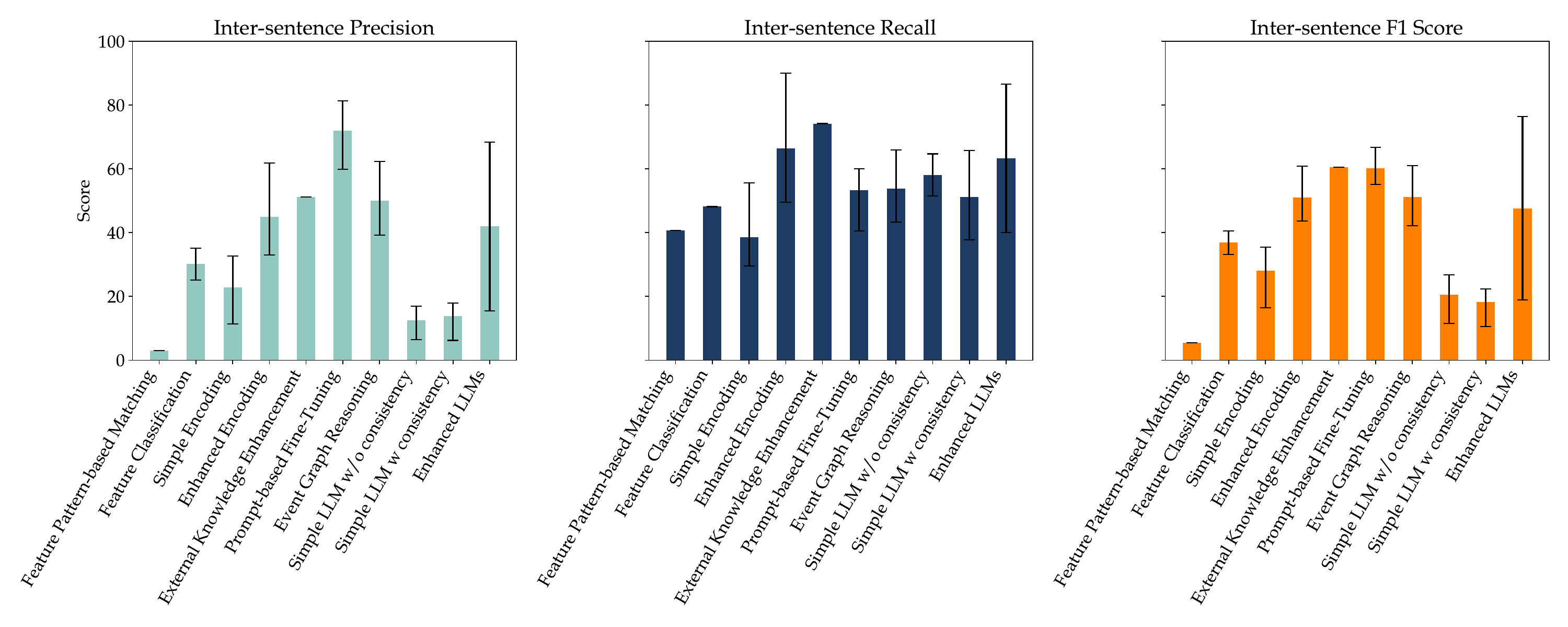}
        \label{fig:esl_inter}
    }
    \caption{Performance comparison (Precision, Recall, F1) on ESL dataset. Error bars represent the range of variability across models.}
    \label{fig:esl}
\end{figure}

{
\textbf{Baselines' Performance on Intra-sentence Task}. \textit{Feature pattern matching-based} methods and \textit{feature classification} methods struggle with balancing precision and recall, resulting in limited overall performance. Shallow feature-based methods, such as DD and VR-C, show high precision but low recall, while simple statistical methods like PPMI underperform in both.
\textit{Deep semantic encoding}-based methods improve recall but produces many false positives, likely due to a limited understanding of "causality" versus "correlation" and the ability of identifying confounding factors (i.e. causal hallucination). Enhanced encoding-based methods show notable improvements, benefiting from richer knowledge and features and better event context representation. Among these approaches, external knowledge enhanced methods show more balanced performance in precision and recall. 
Although \textit{data augmentation}-based methods increase recall over shallow and simple encoding methods, they also introduce more false positives, indicating that the quality of additional training data is crucial for enhancing overall performance.
\textit{Event graph reasoning}, designed for DECI tasks, also performs well on SECI tasks with balanced precision and recall, only slightly behind PLM prompt-tuning and reasoning-enhanced methods. Some models still face challenges with balancing precision and recall due to design nuances.
\textit{Prompts-based fine-tuning} methods achieve a favorable balance between precision and recall, resulting in optimal overall performance. It significantly outperforms other baselines in terms of precision, indicating its capability in suppressing false positives. This advantage is primarily attributed to the effective constraints provided by the prompts.
\textit{Simple LLM}-based methods achieve high recall but with frequent false positives, a result of inherent "causal hallucination" in PLMs/LLMs. The performance of these methods also varies significantly with LLMs' scale. 
The \textit{LLM enhancement} methods, through approaches such as incorporating external knowledge, multi-dimensional reasoning, and self-verification, have effectively mitigated causal hallucinations (improvement of the precision metric). Their performance in zero-shot scenarios has demonstrated the exciting potential of such approaches.}

\begin{table}[htbp]
  \centering
  \caption{{Experimental Results in MAVEN-ERE. All the results have been multiplied by 100, the \textbf{bold} values represent the optimal results, while the \underline{underlined} values indicate the suboptimal results.}}
  \begin{scriptsize}
    \begin{tabular}{|c|c|c|c|c|c|c|c|c|c|c|c|}
         \hline
    \multicolumn{2}{|c|}{\multirow{2}[1]{*}{Type}} & \multirow{2}[1]{*}{Model} & \multicolumn{3}{c|}{Intra-sentence} & \multicolumn{3}{c|}{Inter-sentence} \\
\cline{4-9}    \multicolumn{2}{|c|}{} &       & P     & R     & F1    & P     & R     & F1 \\
    \hline
    \multicolumn{1}{|c|}{\multirow{10}[5]{*}{Deep Semantic Encoding}} & \multirow{5}[2]{*}{Simple Encoding} & LSTM* & 31.5  & 40    & 34.9  & 13.7  & 24.3  & 17 \\
          & \multicolumn{1}{c|}{} & BiLSTM* & 30.1  & 42.5  & 35.2  & 15.3  & 26.8  & 17.9 \\
          & \multicolumn{1}{c|}{} & BERT  & 46.8  & 50.3  & 48.5  & 43    & 46.8  & 44.8 \\
          & \multicolumn{1}{c|}{} & RoBERTa* & 45.9  & 48.8  & 47.2  & 42.3  & 45.1  & 43.6 \\
          & \multicolumn{1}{c|}{} & Longformer* & 39.3  & 55.9  & 44.6  & 35.8  & 46.0  & 39.2 \\
\cline{2-9}          & \multirow{2}[2]{*}{Enhanced Encoding} & SemSIn* & 49.4  & 51.2  & 50.5  & -     & -     & - \\
          & \multicolumn{1}{c|}{} & SemDI* & 45.3  & 51.5  & 48.6  & -     & -     & - \\
\cline{2-9}          & \multirow{3}[2]{*}{DECI Models} & DiffusECI* & 47.5  & 60.3  & 51.7  & 45.5  & 53    & 48.2 \\
          & \multicolumn{1}{c|}{} & KADE* & 58.7  & 64.4  & 60    & 43.7  & 59.3  & 48 \\
          & \multicolumn{1}{c|}{} & SENDIR & 51.4  & 53.6  & 52.5  & 51.9  & 52.8  & 52.4 \\
    \hline
    \multicolumn{2}{|c|}{\multirow{2}[1]{*}{Prompt-based Fine-Tuning}} & ICCL* & 47.5  & 51    & 49.3  & -     & -     & - \\
    \multicolumn{2}{|c|}{} & HOTECI* & 50    & 57.5  & 54.9  & 67.3  & 43.1  & 54.4 \\
    \hline
    \multicolumn{2}{|c|}{\multirow{4}[1]{*}{Event Graph Reasoning}} & ERGO  & 63.1  & 65.3  & \textbf{64.2} & 48.7  & 62    & 54.6 \\
    \multicolumn{2}{|c|}{} & iLIF  & 74.4  & 51.5  & 60.9  & 67.1  & 49.2  & 56.8 \\
    \multicolumn{2}{|c|}{} & PPAT  & 44.6  & 53.7  & 48.0  & 42.8  & 44.5  & 43.3 \\
    \multicolumn{2}{|c|}{} & RichGCN & 41.9  & 50.2  & 45.8  & 35.3  & 47.6  & 40.1 \\
    \hline
    \multicolumn{1}{|c|}{\multirow{10}[3]{*}{LLMs for ECI}} & \multirow{5}[1]{*}{Simple LLM w/o consistency} & LLaMA2-7B* & 12.1  & 50.7  & 19.5  & -     & -     & - \\
          & \multicolumn{1}{c|}{} & text-davinci-002 & 19.6  & 92.9  & 32.4  & -     & -     & - \\
          & \multicolumn{1}{c|}{} & text-davinci-003 & 25.0  & 75.1  & 37.5  & -     & -     & - \\
          & \multicolumn{1}{c|}{} & GPT-3.5-turbo & 19.9  & 85.8  & 32.3  & -     & -     & - \\
          & \multicolumn{1}{c|}{} & GPT-4.0 & 22.5  & 92.4  & 36.2  & -     & -     & - \\
\cline{2-9}          & \multirow{3}[1]{*}{Simple LLM w consistency} & GPT-4o-mini* & -     & -     & -     & 30.8  & 48.3  & 34.4 \\
          & \multicolumn{1}{c|}{} & DeepSeek-Chat* & -     & -     & -     & 37.8  & 53.2  & 40.6 \\
          & \multicolumn{1}{c|}{} & DeepSeek-R1* & -     & -     & -     & 43.7  & 59.2  & 37.8 \\
\cline{2-9}     
          & \multicolumn{2}{c|}{Dr. ECI} & 22.0 & 80.0 & 34.5 & -     & -     & -  \\
          & \multicolumn{2}{c|}{Dr. ECI$^\Delta$} & \textbf{76.6} & \textbf{93.1} & \textbf{84.1} & \textbf{70.7} & \textbf{81.3} & \textbf{75.6} \\
    \hline
    \end{tabular}%
    \end{scriptsize}
  \label{tab:maven}%
\end{table}%

\begin{table}[htbp]
  \centering
  \caption{{Experimental Results in MECI. All the results have been multiplied by 100, the \textbf{bold} values represent the optimal results.}}
  \begin{scriptsize}
        \begin{tabular}{|c|c|c|c|c|c|c|c|c|c|c|c|c|c|c|c|c|}
\hline    \multirow{2}[1]{*}{Type} & \multirow{2}[1]{*}{Model} & \multicolumn{3}{c|}{English} & \multicolumn{3}{c|}{Danish} & \multicolumn{3}{c|}{Spanish} & \multicolumn{3}{c|}{Turkish} & \multicolumn{3}{c|}{Urdu} \\
\cline{3-17}          &       & P     & R     & F1    & P     & R     & F1    & P     & R     & F1    & P     & R     & F1    & P     & R     & F1 \\
    \hline
    {\multirow{6}[2]{*}{\makecell{Deep Semantic \\ Encoding}}} & mBERT & 38.4  & 46    & 41.9  & 25.2  & 26.6  & 25.9  & 43.9  & 41.5  & 42.7  & 36.2  & 48.7  & 41.6  & 31.9  & 34.3  & 33 \\
          & XLMR  & 48.7  & 59.9  & 53.7  & 35.9  & 36.2  & 36    & 50.6  & 49.1  & 49.9  & 44    & 59.4  & 50.5  & 40.4  & 43.2  & 41.8 \\
          & KnowMMR & 42.1  & 45.2  & 43.6  & 43.2  & 32.5  & 37.1  & 39.2  & 49.8  & 43.9  & 34.1  & 14.9  & 20.7  & 44.6  & 25.8  & 32.7 \\
          & SemSIn* & 56.6  & 69.1  & 62.2  & 57.9  & 58.6  & 58.2  & 54.9  & 65.8  & 59.9  & 37.5  & 49.3  & 42.6  & 50.8  & 59.4  & 54.8 \\
          & GCKAN* & 52.8  & 70.5  & 60.4  & 53.5  & 47.3  & 50.2  & 53.8  & 55.4  & 54.6  & 43.2  & 45.8  & 44.5  & 56.1  & 45.9  & 50.5 \\
          & DiffusECI & \textbf{70.1} & 68.3  & \textbf{69.2} & 42.7  & 53.3  & 47.4  & 62.9  & 50.2  & 55.8  & 52.6  & 66.5  & 58.7  & 58.1  & 52.5  & 55.2 \\
    \hline
    {\multirow{3}[1]{*}{\makecell{Event Graph \\ Reasoning}}} & ERGO* & 55.1  & 73.8  & 63.1  & 56.9  & 57.8  & 57.3  & 56.8  & 65.9  & 61    & 35.5  & 43.2  & 39    & 46.9  & 55.6  & 50.9 \\
          & CHEER* & 59.9  & 73.9  & 66.2  & \textbf{61.3} & 62.6  & \textbf{61.9} & 60.2  & 68.8  & 64.2  & 42.9  & 53.2  & 47.5  & 53.8  & 62.3  & 57.7 \\
          & RichGCN & 53.6  & 71.8  & 61.4  & 53.6  & 57.2  & 55.3  & 54.8  & 67.2  & 60.4  & 36.8  & 53.7  & 43.7  & 41    & 60.5  & 48.9 \\
    \hline
    {\multirow{2}[1]{*}{\makecell{Prompt-based \\ Fine-tuning}}} & HOTECI & 66.6  & 67.1  & 66.8  & 50.5  & 63.7  & 56.3  & 60.7  & 60.7  & 60.7  & \textbf{72.5} & \textbf{76.6} & \textbf{74.5} & 59.1  & \textbf{71} & \textbf{64.5} \\
          & KEPT* & 49.5  & 72.6  & 58.9  & 49.3  & 44.6  & 46.8  & 51.2  & 53.8  & 52.5  & 41.6  & 43.2  & 42.4  & 52.8  & 45.1  & 48.6 \\
    \hline
    \multirow{2}[1]{*}{LLM based} & GPT-3.5-turbo & 24.6  & \textbf{79.4} & 37.6  & 10    & \textbf{66.5} & 17.4  & 7.3   & \textbf{74.2} & 13.3  & 27    & 69.3  & 38.8  & 15.7  & 63.8  & 25.2 \\
          & GPT-4o-mini* & 22.3  & 75.9  & 33.9  & 11.3  & 65.4  & 18.7  & 8.6   & 69.7  & 12.5  & 29.7  & 71.6  & 40.5  & 16.2  & 66    & 27.9 \\
    \hline
    {\multirow{2}[1]{*}{Multi-Lingual Model}} & Meta-MK & 55.3  & 71.4  & 62.3  & 36.6  & 47.8  & 41.5  & 57.7  & 61.0  & 59.3  & 58.1  & 66.7  & 62.1  & 39.2  & 61.5  & 47.9 \\
          & GIMC  & 63.4  & 54.8  & 58.8  & 60.2  & 45.2  & 51.6  & \textbf{77.5} & 55.7  & \textbf{64.8} & 70.1  & 60.1  & 64.7  & \textbf{62.1} & 42.4  & 50.4 \\
    \hline
    \end{tabular}%
    \end{scriptsize}
  \label{tab:meci}%
\end{table}%

{
\textbf{Baselines' Performance on Inter-sentence Task}. \textit{Feature pattern-based} methods and \textit{feature classification} methods deliver relatively high recall but with many false positives. \textit{Simple encoding}-based methods show low precision and recall due to difficulty capturing context and causal links, especially between distant events. \textit{Enhanced encoding} methods, whether through text information augmentation or external knowledge augmentation, have effectively mitigated this issue. \textit{Prompt-based fine-tuning} methods again achieve the best-balanced overall performance, particularly excelling in precision by reducing false positives, indicating that high-quality and effective prompts substantially improve PLM understanding of causal patterns. \textit{Event graph reasoning}-based methods also perform well, showing higher recall but slightly lower precision. This drop is likely due to directly modeling events as graph nodes, potentially missing textual details and increasing false positives. However, the method's higher recall results from effectively reducing noise, improving representation quality, and capturing causal links across distant events. \textit{Simple LLM}-based methods perform similarly on DECI and SECI tasks, with high recall but lower precision. However, recall decreases significantly in DECI tasks, suggesting that even advanced PLMs struggle to accurately identify inter-sentence causal relations in unsupervised settings. \textit{Enhanced LLM}-based methods have also significantly mitigated the limitations of simple LLM approaches, demonstrating performance improvements—particularly on the MAVEN-ERE dataset.}

{
\textbf{Baselines' Performance on Multi-lingual ECI Task}. \textit{Prompt-based fine-tuning} method demonstrated relatively stable performance across multi-lingual datasets. \textit{Event graph reasoning}-based model and the model specifically designed for multi-lingual ECI tasks ranked second in performance. \textit{LLM}-based approach underperformed on low-resource language datasets, primarily due to insufficient pre-training corpus coverage for these languages.}

{
\textbf{Cross-domain Generalization Performance}. The cross-domain generalization performance is evaluated on the ESL and MAVEN-ERE datasets, where the training and test sets cover distinct thematic domains, thereby assessing models' ability to generalize across different subject areas. As shown in Figures \ref{tab:ESL} and \ref{tab:maven}, prompt-based fine-tuning models, external knowledge enhanced deep semantic encoding, event graph-based methods, and LLM-enhanced approaches demonstrate superior performance with higher F1 scores. These quantitative results are consistent with the qualitative analysis presented in subsection \ref{perf_assess}, collectively indicating that these methods exhibit stronger robustness when applied to unseen domains.}

\begin{table}[htbp]
\centering
\caption{{Comparison of Event Causality Annotation in ESC, CTB, and MAVEN-ERE Datasets}}
\label{tab:dataset_comparison}
\begin{footnotesize}
\begin{tabular}{|c|c|c|c|}
\hline
{Aspect} & {ESC} & {CTB} & {MAVEN-ERE} \\
\hline
{Relation Types} & 
\makecell{PLOT LINK (loose causal-temporal),\\ Coreference, TLINK} & 
\makecell{CLINK (CAUSE, ENABLE, PREVENT), \\ Temporal relations} & 
\makecell{CAUSE, PRECONDITION, Coreference, \\ Temporal, Subevent} \\
\hline
{Causal Annotation} & 
Explanatory relations, rise/fall actions & 
Explicit causal constructions only & 
Includes causal transitivity \\
\hline
{Annotation Method} & 
\makecell{Expert annotation using CAT \cite{Wilcock2009CAT} tool,\\ Dice coefficient \cite{dicecoef} for IAA \cite{Artstein2017IAA}} & 
\makecell{Rule-based algorithm with \\ manual verification} & 
\makecell{Crowdsourcing with staged \\ annotation and voting} \\
\hline
{Data Characteristics} & 
\makecell{Domain-specific news topics,\\ complex contextual relations} & 
\makecell{Limited to explicit connectives, \\ fewer annotations} & 
\makecell{Broad coverage, integrated \\ event relation network} \\
\hline
\end{tabular}
\end{footnotesize}
\end{table}

\subsection{{Data Annotation Quality and Its Impact}}

{By examining both the datasets and the performance variations of models across different datasets, we conducted a comparative analysis of annotation discrepancies and their effects on model performance across three datasets: CTB, ESL, and MAVEN-ERE. }

\begin{figure}[htbp]
    \centering
    \subfloat[MAVEN-ERE (Intra-Sentence)]{%
        \includegraphics[width=0.85\textwidth]{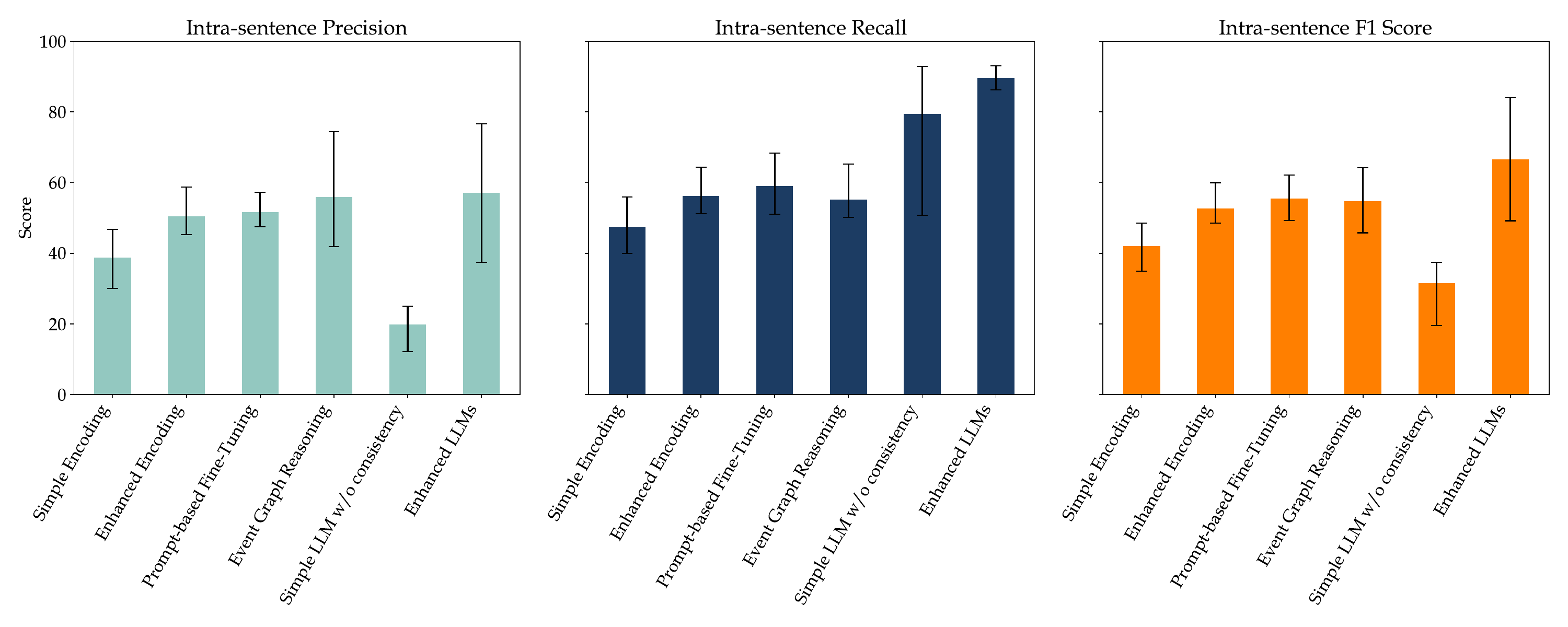}
        \label{fig:maven_intra}
    }\hfill
    \subfloat[MAVEN-ERE (Inter-Sentence)]{%
        \includegraphics[width=0.85\textwidth]{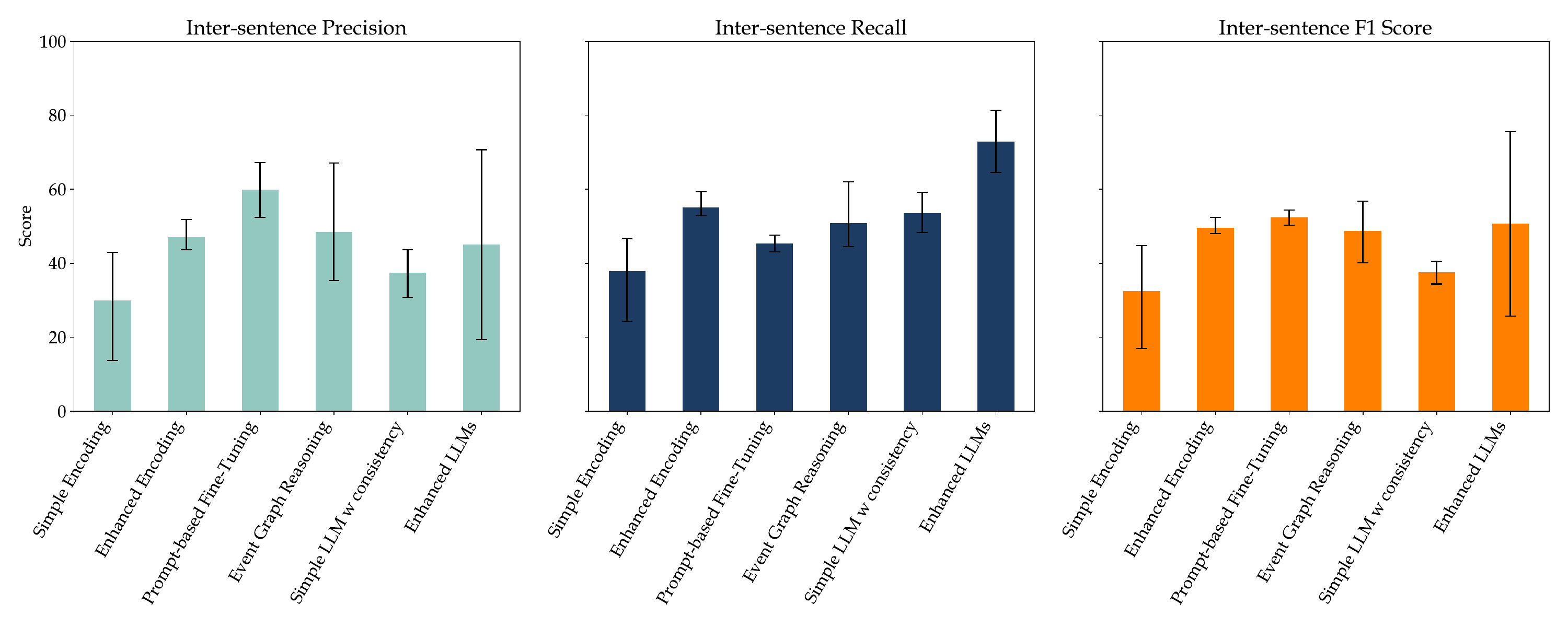}
        \label{fig:maven_inter}
    }
    \caption{{Performance comparison (Precision, Recall, F1) on MAVEN-ERE dataset. Error bars represent the range of variability across models.}}
    \label{fig:maven}
\end{figure}

\begin{figure}[htbp]
    \centering
    \includegraphics[width=0.95\textwidth]{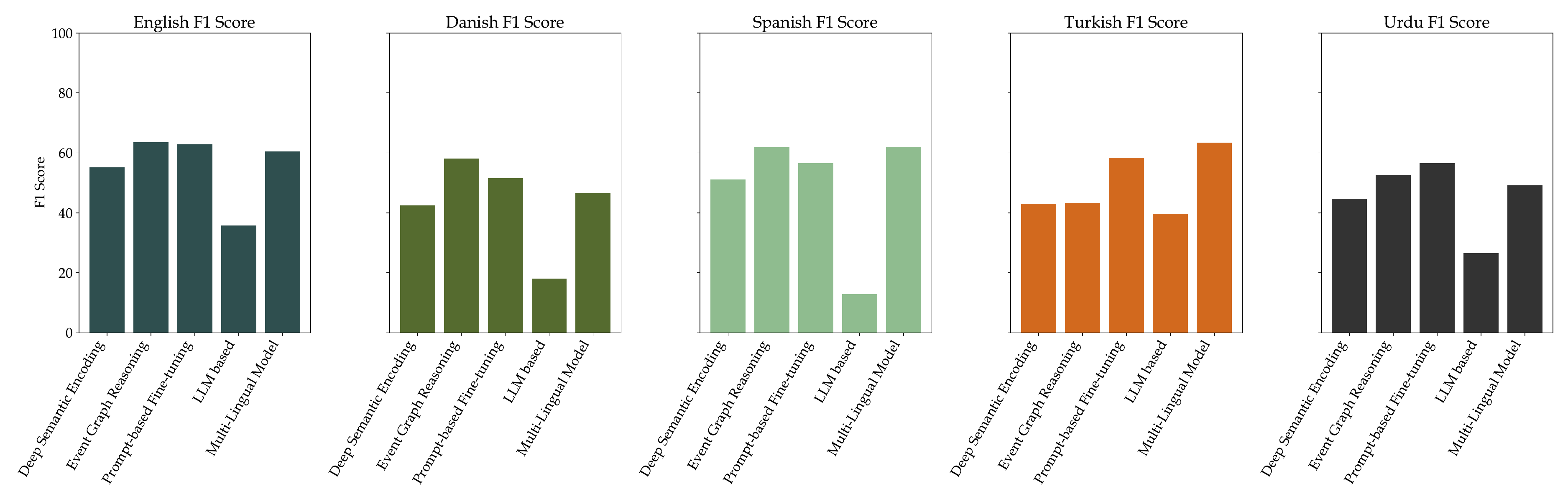}
    \caption{{Performance comparison (Precision, Recall, F1) corss MECI dataset.}}
    \label{fig:meci}
\end{figure}

\begin{figure}[htbp]
    \centering
    \subfloat[(Intra-Sentence)]{%
        \includegraphics[width=0.65\textwidth]{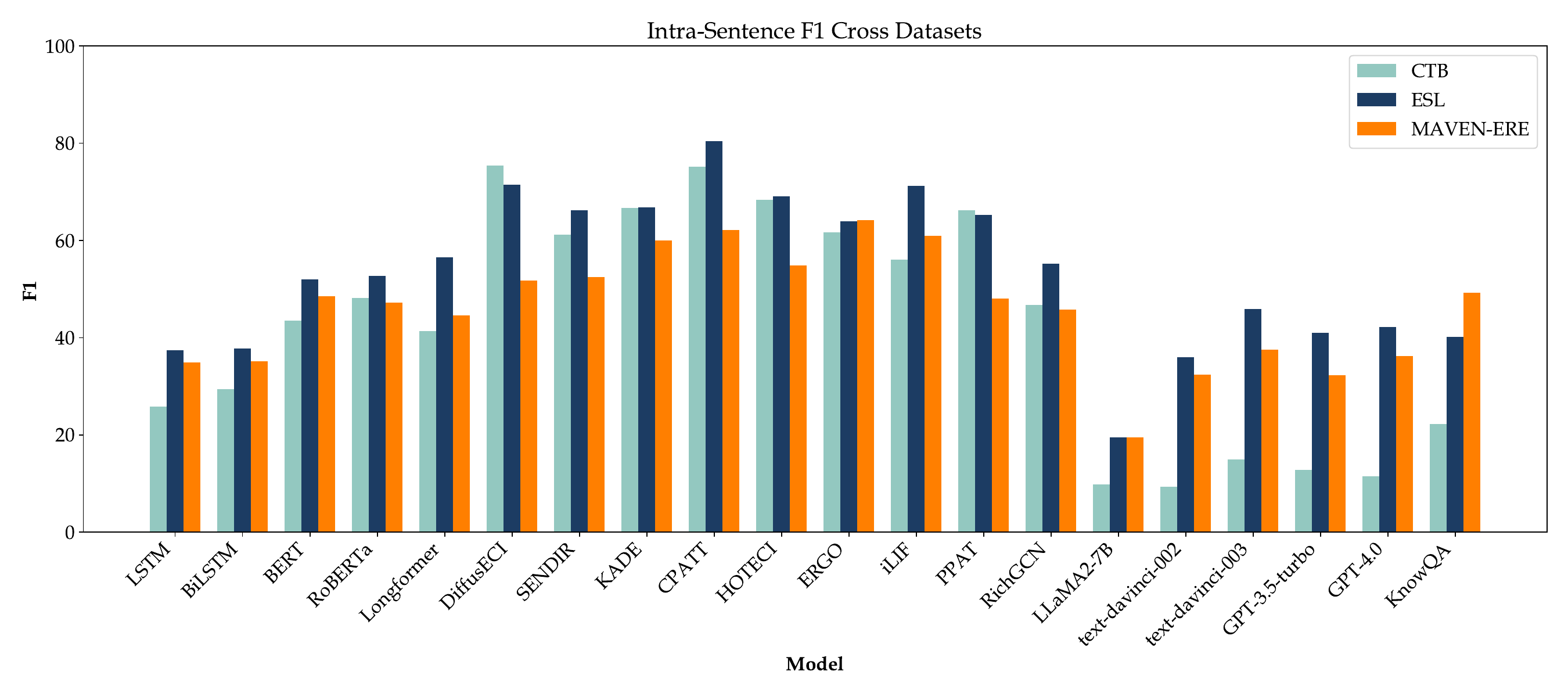}
        \label{fig:intra_diff}
    }\hfill
    \subfloat[(Inter-Sentence)]{%
        \includegraphics[width=0.65\textwidth]{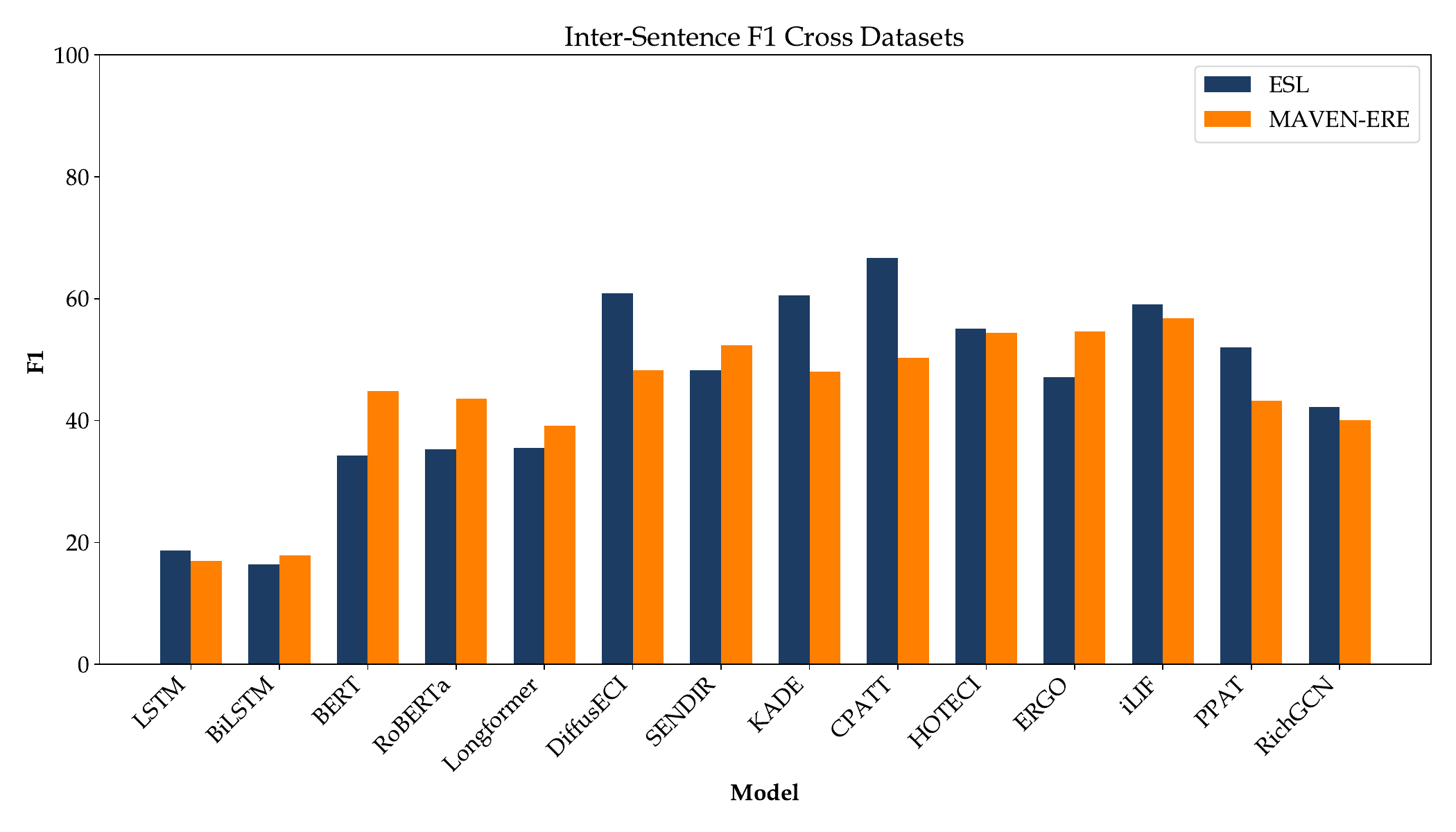}
        \label{fig:inter_diff}
    }
    \caption{{Performance comparison (Precision, Recall, F1) across CTB, ESL, and MAVEN-ERE datasets.}}
    \label{fig:datasets_diff}
\end{figure}

{
\textbf{Analysis of Dataset Discrepancies}. We summarize the main differences among the three datasets in terms of relation types, causal annotation, annotation methods, and data characteristics in Table \ref{tab:dataset_comparison}. Overall, the quality of causal relationship annotations in ESL and MAVEN-ERE is higher than in CTB, with broader coverage of causal relationship types and richer contextual diversity. However, ESL and MAVEN-ERE also have inherent limitations: In ESL, some event causal relationships form directed cycles, which contradicts the basic logic of causality \cite{Pearl2009}. In MAVEN-ERE, automatically inferred causal relationships are not always appropriate, potentially suffering from issues such as scenario drift and threshold effects \cite{reco}.}

{
\textbf{Impact of Data Annotation Quality}. We selected models used across all three datasets in the experiments and compared their performance variations across these datasets, with the results shown in Figure \ref{fig:datasets_diff}. The results highlight that ESL’s annotation quality drives the highest intra-sentence F1 scores, followed by MAVEN-ERE, while CTB lags due to its less robust annotations. However, ESL’s directed cycles and MAVEN-ERE’s inference artifacts reduce inter-sentence performance, indicating that even high-quality annotations face challenges when logical or contextual inconsistencies arise. Advanced models like CPATT and DiffusECI consistently perform well across datasets, suggesting robustness to annotation limitations, while simpler models (e.g., LSTM, LLaMA2-7B) show greater sensitivity to annotation quality, with significantly lower F1 scores on CTB.}

\newpage
\section{{Real-World Applications}}\label{application}
{
ECI has been widely applied in numerous real-world domains. In this section, we review research on ECI applications in healthcare, finance, and social science domains to summarize their current state.}
{
\subsection{ECI in Healthcare}
ECI has become a powerful tool in healthcare, enabling the extraction of cause-effect relationships from diverse textual sources to support clinical decision-making, knowledge discovery, and medical research.
Zhao \cite{zhao2017mining} leverages causalities in diseases, symptoms, treatments, and drug effects from medical literature and clinical data to assist medical diagnosis.
Kabir et al. \cite{kabir2022informative} extracted cause-effect entities from medical literature to support medical information retrieval and knowledge base enrichment.
Gopalakrishnan et al. \cite{medical} extracted causality from Clinical Practice Guidelines for gestational diabetes guidelines. 
Lecu et al. \cite{LECU2024443} identified causality from medical abstracts related to Age-related Macular Degeneration and integrating these relationships into a knowledge graph to enhance medical research and understanding of the disease.
Doan et al. \cite{doan2019extracting} analyzed health-related concerns in daily life by mining causal relationships from tweets about stress, insomnia, and headache.
Hussain et al. \cite{hussain2021practical} convert unstructured clinical text into causal knowledge for supporting healthcare services like clinical decision-making, medical knowledge discovery, and patient persona creation.}
{
\subsection{ECI in Finance}
ECI has become increasingly vital in the finance domain, enabling the extraction of complex cause-effect relationships from unstructured financial texts to support decision-making, event prediction, and the construction of knowledge graphs for investors, regulators, and analysts.
Chen et al. \cite{chen2020pairwise} mined nested causal relationships from financial statements to help investors and regulators understand complex event correlations through a pairwise causality graph approach.
Cao et al. \cite{cao2021cbcp} identified financial causal relationships from unstructured text to help understand connections between financial events and support the construction of event evolutionary graphs.
Nayak et al. \cite{cepn} extracted multiple and overlapping financial causalities from documents like news articles and reports, using a generative approach to overcome limitations of sequence labeling methods.
Sakaji et al. \cite{sakaji2023financial} mined financial causal knowledge from bilingual text data to support expert decision-making by fund managers and financial analysts.
Sun et al. \cite{sun2023leakgan} used one-cause-one-effect relationships for applications in event prediction, question-answering systems, and scenario generation in finance.}
{
\subsection{ECI in Social Science}
ECI plays a transformative role in social science by enabling the automated extraction of cause-effect relationships from diverse textual sources, such as news articles and scholarly papers, to enhance event prediction, theory development, and knowledge synthesis.
Radinsky et al. \cite{radinsky} predict plausible future news events based on current events by mining 150 years of news articles and leveraging large-scale world knowledge ontologies. 
Lei et al. \cite{lei2019event} improve event prediction systems by incorporating context-aware causality relationships between events extracted from texts.
Norouzi et al. \cite{Norouzi_Kleinberg_Vermunt_van_Lissa_2025} systematically identified and extracted causal claims from social science literature to support theory development and methodological rigor in the field.
Chen et al. \cite{chen2020causal} automated the extraction of causal hypotheses and their constituent cause-effect entities from scholarly papers in social sciences to accelerate knowledge synthesis.}

\end{document}